\documentclass[journal,twoside,web]{ieeecolor}
\usepackage{tmi}
\usepackage{cite}
\usepackage{amsmath,amssymb,amsfonts}
\usepackage{algorithmic}
\usepackage{graphicx}
\usepackage{textcomp}
\usepackage[super]{nth}
\usepackage{upgreek}
\usepackage{bm}
\usepackage{array}

\def\BibTeX{{\rm B\kern-.05em{\sc i\kern-.025em b}\kern-.08em
    T\kern-.1667em\lower.7ex\hbox{E}\kern-.125emX}}
\begin{document}
\title{Nonperiodic dynamic CT reconstruction using backward-warping INR with regularization of diffeomorphism (BIRD)}
\author{Muge Du, Zhuozhao Zheng, Wenying Wang, Guotao Quan, Wuliang Shi,
Le Shen, Li Zhang, Liang Li, Yinong Liu, and Yuxiang Xing
\thanks{This work was supported in part by the National Natural Science Foundation of China (Grant No. 62031020) and the grant from Institute of Precision Medicine, Tsinghua University.}
\thanks{Muge Du, Wuliang Shi, Le Shen, Li Zhang, Liang Li, Yinong Liu, and Yuxiang Xing are with the Key Laboratory of Particle \& Radiation Imaging, Department of Engineering Physics, Tsinghua University, Beijing 100084, China. Yuxiang Xing is the corresponding author (e-mail:xingyx@mail.tsinghua.edu.cn).}
\thanks{Zhuozhao Zheng is with Department of Radiology, Beijing Tsinghua Changgung Hospital, School of Clinical Medicine, Tsinghua Medicine, Tsinghua University, Beijing, 102218, China (e-mail: zzza00509@btch.edu.cn, ORCID: 0000-0001-8547-6670).}
\thanks{Wenying Wang is with United Imaging  Healthcare, North America (e-mail: wenying.wang@united-imaging.com).}
\thanks{Guotao Quan is with Shanghai United Imaging Healthcare Co., Ltd., Shanghai 201815, China (e-mail: guotao.quan@united-imaging.com).}}
\maketitle

\begin{abstract}
Dynamic computed tomography (CT) reconstruction faces significant challenges in addressing motion artifacts, particularly for nonperiodic rapid movements such as cardiac imaging with fast heart rates. While existing methods effectively handle periodic motion, they struggle with the extreme limited-angle problems inherent in nonperiodic cases. Deep learning methods have improved performance but face generalization challenges with complex structures and nonperiodic motion. Recent implicit neural representation (INR) techniques show promise through self-supervised deep learning, but face critical limitations: computational inefficiency due to forward-warping modeling, difficulty balancing DVF complexity with anatomical plausibility, and challenges in preserving fine details without additional patient-specific pre-scans. This paper presents a novel INR-based framework, BIRD, for nonperiodic dynamic CT reconstruction that addresses these challenges through four key contributions:  (1) backward-warping deformation modeling that enables direct computation of each dynamic voxel with significantly reduced computational demands, (2) diffeomorphism-based DVF regularization that ensures anatomically plausible deformations while maintaining representational capacity for complex motion patterns, (3) motion-compensated analytical reconstruction that enhances fine details while providing implicit supervision for DVF estimation without requiring additional pre-scans, and (4) dimensional-reduction design for efficient 4D coordinate encoding. Through various simulations and practical studies, including digital and physical phantoms and retrospective patient data, we demonstrate the effectiveness of our approach for nonperiodic dynamic CT reconstruction with enhanced details and reduced motion artifacts. The proposed framework enables more accurate dynamic CT reconstruction with potential clinical applications, such as one-beat cardiac reconstruction, cinematic image sequences for functional imaging, and motion artifact reduction in conventional CT scans affected by failed breath-holding or other movements.
\end{abstract}

\begin{IEEEkeywords}
Computed tomography, deep learning, implicit neural representation, motion artifact reduction.
\end{IEEEkeywords}

\section{Introduction}
\label{sec:introduction}
\IEEEPARstart{D}{ynamic} computed tomography (CT) reconstruction is critical for medical applications ranging from cancer treatment to cardiovascular disease diagnosis. Unlike normal CT reconstruction, dynamic CT reconstruction needs to address the motion artifacts that cause blurring and ghosting. While dynamic images with relatively slow motion can be effectively reconstructed using established methods, the rapid motion cases, such as cardiac imaging with fast heart rate, remain challenging.

Dynamic CT reconstruction approaches differ based on motion characteristics. For periodic motion, a common approach involves scanning across multiple cycles and binning projections into distinct phases according to motion signals. This binning can be either retrospective or prospective, with the primary objective of subsequent reconstruction being to mitigate sparse-view artifacts and residual motion artifacts within each phase. However, these approaches often fail for patients with arrhythmia, producing blurred reconstructions due to inter-cycle inconsistencies. For nonperiodic motion, such as the imaging objects using single-cycle or "one-beat" scanning techniques, the inter-cycle inconsistency is addressed, but the severer limited-angle reconstruction presents significant challenge, which is the main problem this work attempts to address.

Traditional dynamic CT reconstruction methods exploit spatial-temporal correlations in 4D images through two principal approaches: regularization-based techniques and deformation-based modeling. The former employs regularization terms such as spatial-temporal total variation \cite{ritschlIterative4DCardiac2012}, temporal non-local means \cite{jiaFourdimensionalConeBeam2012}, tensor framelet \cite{gao4DConeBeam2012}, and prior image constraint \cite{chenPriorImageConstrained2008} to regularize per-phase reconstruction. While effective for objects exhibiting mild motion, these methods face limitations when handling strong motion due to insufficient motion modeling. The latter approach assumes that phase images originate from a shared static reference image via deformation vector fields (DVFs), focusing on the estimation of both DVFs and reference images. This estimation can be performed either separately \cite{ritOntheflyMotioncompensatedConebeam2009, parkMotionmapConstrainedImage2013} or within unified frameworks \cite{wangSimultaneousMotionEstimation2013, zhouGeneralSimultaneousMotion2021}. Some methods \cite{hahnReductionMotionArtifacts2016, hahnMotionCompensationRegion2017, lebedevMotionCompensationAortic2022} estimate DVFs from partial angle reconstruction (PAR) images, subsequently applying these DVFs to deform PARs before summation to generate a motion-compensated reconstruction. When historical scan data is available, prior images can be deformed by estimated DVFs to match the projection data \cite{wangHighqualityFourdimensionalConebeam2013}, and historical DVFs and images can serve as additional constraints \cite{zhangTechniqueEstimating4DCBCT2013, moryMotionawareTemporalRegularization2016}. Although demonstrating efficacy for periodic motion or simple non-periodic cases, these methods show reduced performance in complex non-periodic scenarios, primarily due to their inability to address extreme limited-angle reconstruction challenges.

In recent years, deep learning-based methods have outperformed traditional dynamic CT reconstruction.  Similar to traditional approaches, current deep learning-based methods can also be broadly categorized into two paradigms. The first employs deep learning to mitigate the artifacts in phase images, often exploiting inter-phase redundancy, such as extracting complementary information from phase-averaged images or neighboring phases \cite{madestaSelfcontainedDeepLearningbased2020, chen4DAirNetTemporallyresolvedCBCT2020,jiangEnhancement4DConeBeam2022, dengTTUNetTemporal2023a}.  The second integrates deformation-based modeling with deep learning, either through refinement of DVFs \cite{huangUnetbasedDeformationVector2020, tengRespiratoryDeformationRegistration2021, wangDeepOrganNetOntheFlyReconstruction2020, chenEstimateCompensateHead2024} and images \cite{guoSpatiotemporalVolumetricInterpolation2020, zhangDeepLearningbasedMotion2022, yangFourDimensionalConeBeam2023}, or by learning the static image-DVF relationship in feature spaces \cite{jiangFastFourdimensionalConebeam2022}. Despite their success, However, these methods may exhibit limited generalization when applied to unseen cases with complex anatomical structures or motion patterns, due to the limitation of training dataset diversity.

A recent advancement in self-supervised deep-learning, implicit neural representation (INR) \cite{mildenhallNeRFRepresentingScenes2020a}, has shown great potential for ill-posed CT reconstruction \cite{sunCoILCoordinateBasedInternal2021a, zangIntraTomoSelfsupervisedLearningbased2021, shenNeRPImplicitNeural2021, zhaNAFNeuralAttenuation2022, corona-figueroaMedNeRFMedicalNeural2022}. INR-based approaches parameterize image through neural networks and perform reconstruction via self-supervised optimization, minimizing the discrepancy between predicted and measured projection data. This methodology outperforms conventional optimization-based techniques by introducing nonlinear and implicit image constraints while avoiding generalization issues through single-case optimization. For dynamic CT reconstruction, several INR extensions of deformation-based modeling have shown encouraging results, particularly for non-periodic motion scenarios \cite{reedDynamicCTReconstruction2021c, zhangDynamicConebeamCT2023a, shaoDynamicCBCTImaging2024}. These implementations typically employ INR-based methods face to represent static reference images, while modeling DVFs through hybrid model that combining INR with polynomial \cite{reedDynamicCTReconstruction2021c}, PCA \cite{zhangDynamicConebeamCT2023a}, and B-spline \cite{shaoDynamicCBCTImaging2024}. Among these methods, \cite{zhangDynamicConebeamCT2023a} incorporates patient-specific prior DVFs and images from historical 4D-CT scans as supplementary information source, while \cite{reedDynamicCTReconstruction2021c, shaoDynamicCBCTImaging2024} achieve prior-model-free reconstruction. Nevertheless, existing INR-based methods still present several limitations:

1) Most existing methods employ forward-warping deformation modeling \cite{reedDynamicCTReconstruction2021c, zhangDynamicConebeamCT2023a, shaoDynamicCBCTImaging2024}, wherein static-to-dynamic forward DVFs and a static reference image are independently represented by INRs. Since dynamic coordinates cannot be directly mapped to their static counterparts through analytical inversion, the reconstruction of even a single dynamic voxel necessitates computing the entire DVF across the full spatial domain and multiple static image queries. This leads to high computation and memory costs scaled with increasing spatial and temporal resolution during INR training. Consequently, these INR-based methods become impractical for high-resolution dynamic image reconstruction. 

2) Using INR to represent DVF faces a challenge in balancing representational capacity with anatomical plausibility. Existing methods use hybrid representation, combining INR with explicit functions, such as polynomial \cite{reedDynamicCTReconstruction2021c}, PCA \cite{zhangDynamicConebeamCT2023a}, and B-spline \cite{shaoDynamicCBCTImaging2024}, producing smooth and continuous DVF estimation, though potentially at the cost of limited representational capacity for complex deformations. In contrast, pure INR-based approaches with advanced encoding methods, e.g., multi-resolution hashgrid encoding \cite{mullerInstantNeuralGraphics2022}, can capture intricate motion patterns, but frequently yield irregular, noisy, and anatomically unrealistic results. This is due to the insufficient DVF regularization, a critical factor in maintaining the delicate balance between complexity and anatomical-realism. The challenge is particularly acute in extreme limited-angle extreme limited-angle scenarios for nonperiodic dynamic CT reconstruction.

3) INR-based methods typically suffer from over-smoothness and require extensive training iterations to capture fine details. While some methods incorporate prior information from additional patient-specific pre-scans to enhance DVF or image estimation \cite{zhangDynamicConebeamCT2023a}, this is impractical in clinical scenarios where pre-scans are unavailable or insufficiently representative of current anatomical states. Ideally, high-frequency details should be extracted directly from the current CT projection data through analytical reconstruction. However, conventional analytical reconstruction of dynamic CT projections introduces significant motion artifacts that degrade image quality—precisely why more sophisticated approaches such as INRs are needed. This creates a paradoxical challenge: INRs are required to address motion artifacts, yet themselves struggle to preserve fine details without enhancement from the original projection data that is compromised by motion. Breaking this circular dependency to effectively utilize dynamic projection data for detail enhancement in INR-based reconstruction remains a significant challenge.

To address these challenges of nonperiodic dynamic CT reconstruction and the limitations of existing INR-based methods, we propose a novel framework, BIRD, that leverages backward-warping INR with diffeomorphism regularization. The main contributions are as follows:

1) To overcome the limitation of existing forward-warping dynamic INRs in high-resolution reconstruction, namely their excessive computational demands and limited representational capacity, we adopt backward-warping deformation for dynamic CT image modeling. Unlike forward-warping methods that require whole-image DVF computation and multiple static image queries per voxel, our backward-warping approach enables direct computation of each dynamic voxel by one DVF query and one static image query only, significantly reducing computational and memory requirements for ray-based training.

2) To address the challenge of balancing complexity and anatomical realism in DVF estimation, we propose a DVF regularization method that encourages approximately diffeomorphism, through bidirectional DVF estimation and an inverse-consistency-based diffeomorphism regularization loss. This ensures DVFs to be continuous and realistic while maintaining the representation capability for complex motion patterns, and also enhances robustness in ill-posed dynamic CT reconstruction.

3) To solve the challenge of extracting high-frequency information directly from the dynamic projection data, we propose a novel approach that combining INR-based DVF estimation with analytical reconstruction, which is used for detail-enhancement in the final reconstruction.

4) To more effectively address the ill-posed problem, we propose a dimensional-reduction design for the INR's encoding block that efficiently processes 4D input coordinates within our framework.

The rest of the paper is organized as follows: Section \ref{sec:methods} introduces the methodology, including basic materials, the proposed framework, the dimensional-reduction design for 4D encoding, and the implementation details. Section \ref{sec:results} presents the simulation and practical studies. Section \ref{sec:conclusion} is the conclusion and discussion.

\section{Methods}
\label{sec:methods}

\subsection{Basic materials}
\subsubsection{Mathematical model of Dynamic CT imaging}
\label{sec:methods-basic-math}

Let $\mu\left( {\mathbf{x},t} \right)$ be the attenuation map of a scanned dynamic object at spatial location $\mathbf{x}$ and time $t$, $\mathbf{p} \in \mathbb{R}^{N_{U}N_{V} \times 1}$ be a vector of the post-log X-ray CT projection of the dynamic object with $N_{U}$ detector elements and $N_{V}$ views (frames).

According to the forward model of X-ray CT imaging, each element of $\mathbf{p}$ is a line integration of $\mu\left( {\mathbf{x},t} \right)$ along the ray intersection path:

\begin{equation}
  p_{u,v} = {\int_{l_{u,v}}^{~}{\mu\left( {\mathbf{x},t_{v}} \right)d\mathbf{x}}}
  \label{eq:math-proj}
\end{equation}

where $l_{u,v}$ is the ray path with $u = 1,\ldots,N_{U}$ indexing detector elements and $v = 1,\ldots,N_{V}$ indexing views, and $t_{v}$ is the corresponding time of the $v^\mathrm{th}$ view. The temporal sample number equals to the view number.

CT reconstruction usually uses a discretized version of forward model:

\begin{equation}
  p_{u,v} = {\sum\limits_{n = 1}^{N}{\mu\left( {\mathbf{x}_{n},t_{v}} \right)H\left( {u,v,n} \right)}}
  \label{eq:math-proj-discr}
\end{equation}

where $\mathbf{x}_{n}$ is the voxel coordinate, and $H\left( {u,v,n} \right)$ is the weight denoting the contribution of the $n^\mathrm{th}$ voxel to the $u^\mathrm{th}$ detector bin at the $v^\mathrm{th}$ view.

In matrix-vector format, \eqref{eq:math-proj-discr} becomes:

\begin{equation}
  \left\lbrack \mathbf{p} \right\rbrack_{,v} = \mathbf{H}_{v}\bm{\upmu}\left( t_{v} \right),~\forall v \in \left\lbrack 1,~N_{V} \right\rbrack
  \label{eq:math-proj-matvec}
\end{equation}

with $\bm{\upmu} \in \mathbb{R}^{N \times 1}$ being an image at moment $t_v$ and $\mathbf{H}_{v} \in \mathbb{R}^{N_{U} \times N}$ being the system matrix for the $v^\mathrm{th}$ view.

Given a projection $p_{u,v}$, the reconstruction of $\bm{\upmu}$ can be formulated as a weighted least square problem under Gaussian noise:

\begin{equation}
  \hat{\bm{\upmu}}\left( t_{v} \right) = 
  \mathop{\arg\min}_{\bm{\upmu}\left( t_{v} \right)} 
  \left\| {\mathbf{H}_{v}\bm{\upmu}\left( t_{v} \right) - \left\lbrack \mathbf{p} \right\rbrack_{,v}} \right\|_{\mathbf{\Sigma}}^{2}~,~\forall v \in \left\lbrack 1,~N_{V} \right\rbrack
  \label{eq:math-wls}
\end{equation}

with $\mathbf{\Sigma}$ denoting the noise variance in projection data. Since each $\hat{\bm{\upmu}}\left( t_{v} \right)$ is only constrained by one view of projection, it’s an extreme case of limited angle problem so that impossible to find satisfying solutions for \eqref{eq:math-wls} directly. To address this issue, some methods ignore high frequency motion and use projections covering multiple views to mitigate the data-incompleteness, i.e., reconstructing a time-average image across a time window. Mathematically, a time-average reconstruction is obtained by:

\begin{equation}
  \hat{\left\langle \bm{\upmu} \right\rangle}_{t_{v}\sim t_{v'}} =
  \mathop{\arg\min}_{{\langle\bm{\upmu}\rangle}_{t_{v}\sim t_{v'}}}
  \left\| {\begin{bmatrix}
  {\mathbf{H}_{v}\left\langle \bm{\upmu} \right\rangle_{t_{v}\sim t_{v'}}} \\
  \ldots \\
  {\mathbf{H}_{v'}\left\langle \bm{\upmu} \right\rangle_{t_{v}\sim t_{v'}}}
  \end{bmatrix} - \begin{bmatrix}
  \left\lbrack \mathbf{p} \right\rbrack_{,v} \\
  \ldots \\
  \left\lbrack \mathbf{p} \right\rbrack_{,v'}
  \end{bmatrix}} \right\|_{\mathbf{\Sigma}}^{2}
  \label{eq:math-wls-time-avg}
\end{equation}

where $\hat{\left\langle \bm{\upmu} \right\rangle}_{t_{v}\sim t_{v'}}$ denotes the time-average of dynamic images within the time window of $t_{v}\sim t_{v'}$. In this case, the nonperiodic dynamic CT reconstruction becomes a trade-off controlled by the length of time window $\mathrm{\Delta}t = t_{v'} - t_{v}$, where a large $\mathrm{\Delta}t$ reduces the limited-angle artifact, but sacrifices the temporal resolution.

To further overcome the ill-posed limited-angle problem, the spatial-temporal relationship of the non-periodic dynamic CT image have been exploited. This leads to methods including additional regularization terms, deformation-based image modeling of $\hat{\bm{\upmu}}$, and incorporation of deep-learning. Despite the mitigation of ill-poseness, using \eqref{eq:math-wls-time-avg} can lead to inadequate exploration of the valuable correlation among all projection data.

\subsubsection{Implicit Neural Representation}
\label{sec:methods-basic-inr}

Implicit Neural Representation (INR), originated from NeRF \cite{mildenhallNeRFRepresentingScenes2020a}, is a special category of neural network-based methods that represents a continuous field as the function mapping from input coordinates to field values. Unlike explicit representations that treat a field as discrete vectors/tensors, INR uses neural networks to parameterize a continuous field $f: \mathbb{R}^n \rightarrow \mathbb{R}^m$, which maps from an n-dimensional coordinate $\mathbf{x} \in \mathbb{R}^n$ space to an m-dimensional signal space, the general formulation of INR to represent the field is:

\begin{equation}
  \hat{f}(\mathbf{x}) = \phi(e(\mathbf{x})).
  \label{eq:inr-general}
\end{equation}

where $e: \mathbb{R}^n \rightarrow \mathbb{R}^q$  is an encoding block that transforms the coordinates into feature vectors,  $\phi: \mathbb{R}^q \rightarrow \mathbb{R}^m$ is a prediction block that maps the encoded features into field value estimation. A typical choice of $e$ is multi-resolution hashgrid encoding \cite{mullerInstantNeuralGraphics2022}. The prediction block $\phi$ is usually based on Multi-Layer Perceptron (MLP).

\subsubsection{Forward-warping v.s. backward-warping for dynamic CT reconstruction}
\label{sec:methods-basic-warping}

In deformation-based modeling of dynamic CT image, each temporal frame is represented by deforming a shared static reference image via a time-dependent deformation vector field (DVF). Two fundamental deformation types are forward-warping and backward-warping, each with distinct computational implications for dynamic CT reconstruction.

In forward warping, each static spatial coordinate $\mathbf{x}_\mathrm{s}$ is mapped to a dynamic spatial coordinate $\mathbf{x}_\mathrm{d}$ at time $t$ by a DVF $\mathbf{D}_{\mathrm{s} \rightarrow \mathrm{d}}$:

\begin{equation}
  \mathbf{x}_\mathrm{d} = \mathbf{x}_\mathrm{s} + \mathbf{D}_{\mathrm{s} \rightarrow \mathrm{d}}\left( {\mathbf{x}_\mathrm{s},t} \right)
  \label{eq:forward-warping}
\end{equation}

Conversely, in backward-warping deformation, each dynamic coordinate $\mathbf{x}_\mathrm{d}$ at time $t$ is mapped into a static coordinate $\mathbf{x}_\mathrm{s}$ by a DVF $\mathbf{D}_{\mathrm{d} \rightarrow \mathrm{s}}$:

\begin{equation}
  \mathbf{x}_\mathrm{s} = \mathbf{x}_\mathrm{d} + \mathbf{D}_{\mathrm{d} \rightarrow \mathrm{s}}\left( {\mathbf{x}_\mathrm{d},t} \right)
  \label{eq:backward-warping}
\end{equation}

The choice of deformation approach substantially impacts computational and memory requirements in the reconstruction of dynamic images:

For forward-warping, reconstruction involves a two-step process. Firstly, since $\mathbf{x}_\mathrm{s}$ cannot be analytically determined from $(\mathbf{x}_\mathrm{d},t)$ in \eqref{eq:forward-warping} due to the non-invertible nature of the mapping, a complete static-to-dynamic warp field must be computed. This includes evaluating $\mathbf{D}_{\mathrm{s} \rightarrow \mathrm{d}}$ at time $t$ across all spatial locations. Secondly, this warp field is applied to transform the static image to the dynamic image, requiring interpolation among multiple pixels of the static image to determine the value at each $(\mathbf{x}_\mathrm{d},t)$. Consequently, reconstructing even a single voxel necessitates whole-image evaluation of the DVF and multiple queries to the static image, resulting in significant computational and memory overhead. 

In contrast, backward-warping offers a more efficient reconstruction process. Given $(\mathbf{x}_\mathrm{d},t)$, the corresponding static coordinate $\mathbf{x}_\mathrm{s}$ can be directly computed using \eqref{eq:backward-warping}, requiring only a single evaluation of the DVF. Subsequently, the dynamic image value at $(\mathbf{x}_\mathrm{d},t)$ is obtained by a single query to the static image at $\mathbf{x}_\mathrm{s}$, which substantially reduces computation. Thus, for applications requiring high spatial and temporal resolution reconstruction, backward-warping is a significantly more efficient approach.

\subsubsection{Notations}

The notations used in the paper are summarized in Table-\ref{tab:notation}.

\begin{table}
\caption{Summary of notations}
\setlength{\tabcolsep}{3pt}
\begin{tabular}{|p{30pt}|p{60pt}|p{135pt}|}
\hline\hline
Notation& 
Position& 
Description \\
\hline
$\mu$& 
plain, superscript& 
attenuation image of CT-scanned object; CT image INR \\

$\mathbf{p}$& 
plain& 
post-log CT projection \\

$\mathbf{x}$& 
plain& 
3D spatial coordinate \\

$t$& 
plain& 
1D spatial coordinate \\

$u$& 
plain, subscript& 
index of detector elements \\

$v$& 
plain, subscript& 
index of projection views \\

$l$& 
plain& 
X-ray intersection path \\

$\mathbf{H}$& 
plain& 
weight matrix of a discretized version of CT forward model \\

$\mathbf{\Sigma}$& 
subscript& 
noise variance in projection data \\

$\hat{\cdot}$& 
plain& 
predicted \\

$\left\langle {\cdot} \right\rangle$& 
plain& 
time-average operation \\

$f$& 
plain& 
field function \\

$e$& 
plain& 
INR’s Encoding block \\

$\phi$& 
plain& 
INR’s Prediction block \\

$\mathbf{D}$& 
plain, superscript& 
deformation vector field (DVF); DVF INR \\

$\mathrm{PAR}$& 
plain, superscript& 
partial angle reconstruction \\

$\mathrm{AR}$& 
plain, subscript& 
analytical reconstruction \\

$K$& 
plain& 
number of PARs \\

$\mathrm{d}$& 
subscript& 
dynamic \\

$\mathrm{s}$& 
subscript& 
static \\

$\rightarrow$& 
subscript& 
warping \\

$\mathbf{F}$& 
plain& 
the feature used for the dual-feature description of general dynamic objects\\

$\mathrm{TP}$& 
subscript& 
topology-preserving \\

$\mathrm{FF}$& 
subscript& 
free-form \\

$\mathrm{4D}$& 
subscript& 
4D spatial-temporal space \\

$\rho^\mathrm{FD}$& 
plain& 
Fourier-domain distance \\

$\mathcal{L}$& 
plain& 
loss function \\

$\mathrm{FDL}$& 
subscript& 
fidelity \\

$\mathrm{DM}$& 
subscript& 
diffeomorphism regularization \\

$\mathrm{RGT}$& 
subscript& 
registration \\

$\lambda$& 
plain& 
loss weight \\

$\nabla$& 
plain& 
Jacobian \\

$\left\| \cdot \right\|_{F}$& 
plain& 
Frobenius norm \\

$\otimes$& 
plain& 
element-wise multiplication \\

$\gamma$& 
plain& 
multi-resolution hashgrid encoding function \\

$\varphi$& 
plain& 
single-hidden-layer MLP \\

$h$& 
plain& 
hierarchical aligned feature \\

\hline\hline
\end{tabular}
\label{tab:notation}
\end{table}

\subsection{Dynamic CT reconstruction using backward-warping INR with regularization of diffeomorphism (BIRD)}

In this work, we propose a novel framework for nonperiodic dynamic CT reconstruction based on implicit neural representations (INRs). We describe a general dynamic object by two complementary features: a Topology-Preserving feature ($\mathbf{F}_\mathrm{TP}$) and a Free-Form feature ($\mathbf{F}_\mathrm{FF}$). A backward-warping model is used to  represent topology-preserving motion in the $\mathbf{F}_\mathrm{TP}$. The $\mathbf{F}_\mathrm{FF}$ takes a hyper-space feature to capture more general dynamic variations, including topological and intensity changes. This dual representation overcomes the limitations of purely deformation-based methods, enabling reconstruction of a wider range of dynamic objects with topological changes such as air bubble splitting/merging, tumor evolution, and intensity variations such as in CT angiography and CT perfusion.

The framework of BIRD consists of three collaborative modules: 1) a Bidirectional DVF Estimation Module that establishes coordinate alignment across raw dynamic data  and intermediate static image space; 2) an Analytical Reconstruction Module that focuses on detail preservation from projections, with motion-compensation facilitated by the estimated DVF in module 1; 3) an INR-based Reconstruction Module that estimates both topology-preserving and free-form features of a dynamic CT image.

\begin{figure}[!t]
\centerline{\includegraphics[width=\columnwidth]{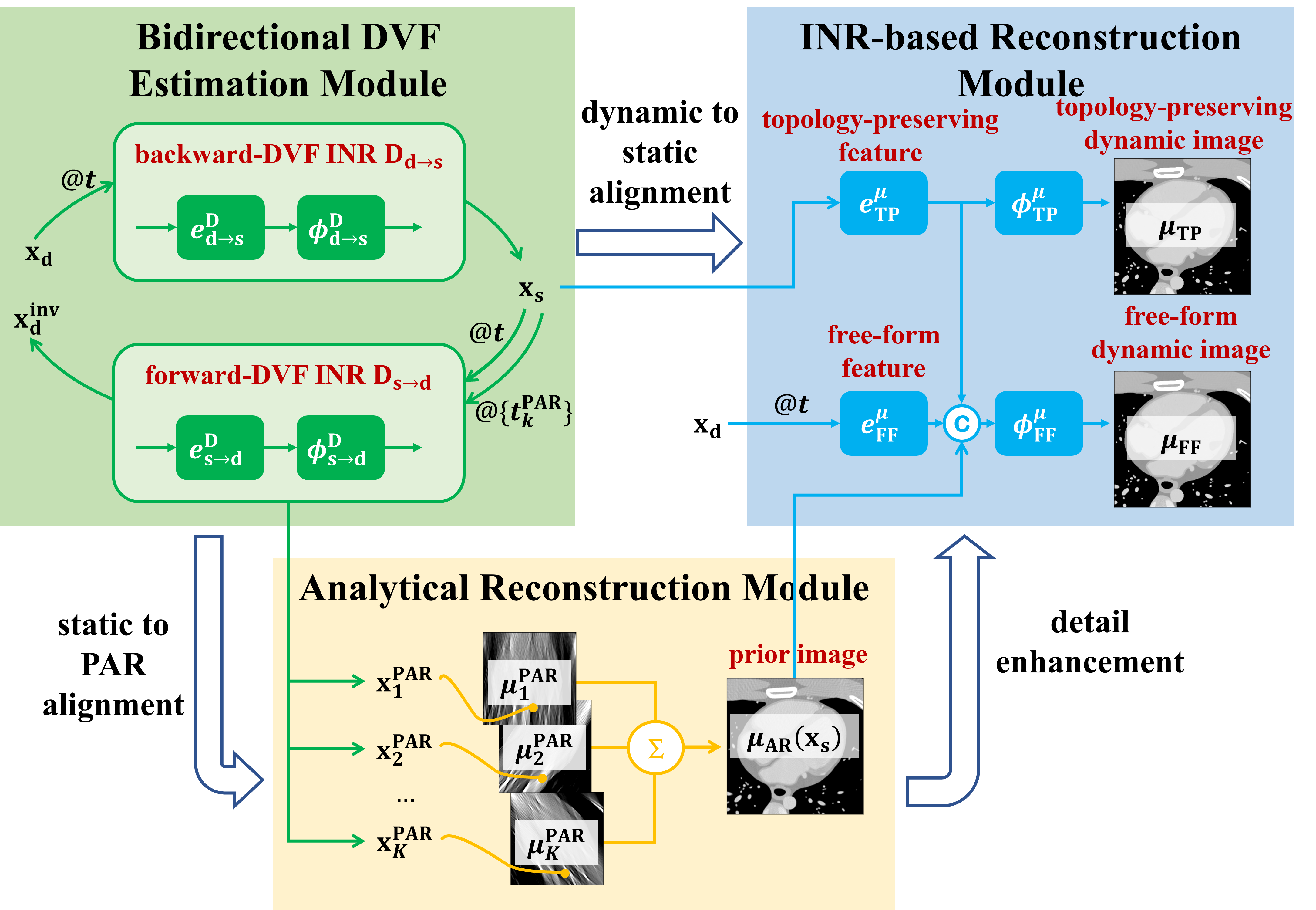}}
\caption{Illustration of the proposed framework.}
\label{fig-framework}
\end{figure}

The overall framework is illustrated in Figure \ref{fig-framework}. The following subsections provide detailed descriptions of the framework.

\subsubsection{Dual-feature description of general dynamic objects}
\label{sec:methods-modules-dualfeature}

We describe a general dynamic object by a topology-preserving feature and a free-form feature. The two features and corresponding image reconstruction are predicted in the INR-based Image Reconstruction Module, which receives the needed information provided by other two modules.

To compute the topology-preserving feature at a set of dynamic coordinates, we first use backward-warping deformation to compute the static coordinates, and then encoded the static coordinates to feature vectors. We use backward-warping for its efficiency as analyzed in Section II-A-3 in high resolution reconstruction. Given a dynamic coordinate $(\mathbf{x}_\mathrm{d},t)$, the static coordinate $\mathbf{x}_\mathrm{s}$ is computed using \eqref{eq:backward-warping}. Next, the static coordinate $\mathbf{x}_\mathrm{s}$ is encoded using an encoding block $e_\mathrm{TP}^{\mu}$ in the INR-based Reconstruction Module:

\begin{equation}
  \mathbf{F}_\mathrm{TP} = e_\mathrm{TP}^{\mu}\left( \mathbf{x}_{s} \right)
  \label{eq:c-tp}
\end{equation}

$\mathbf{F}_\mathrm{TP}$ is the topology-preserving feature. A topology-preserving dynamic image $\mu_\mathrm{TP}$ is thereby reconstructed based on $\mathbf{F}_\mathrm{TP}$:

\begin{equation}
  \mu_\mathrm{TP}\left( {\mathbf{x}_\mathrm{d}, t} \right) = \phi_\mathrm{TP}^{\mu}\left( \mathbf{F}_\mathrm{TP} \right) = \phi_\mathrm{TP}^{\mu}\left( {e_\mathrm{TP}^{\mu}\left( \mathbf{x}_\mathrm{s} \right)} \right)
  \label{eq:mu-tp}
\end{equation}

where $\phi_\mathrm{TP}^{\mu}$ is a prediction block in the INR-based Reconstruction Module.

Using the topology-preserving feature alone can only model topology-invariant motion. Inspired by HyperNeRF \cite{parkHyperNeRFHigherdimensionalRepresentation2021}, we introduce a Free-Form Feature to extend the representation into a hyper-space, allowing the modeling of more general dynamic changes. The free-form feature is derived by directly transforming the dynamic 4D coordinates to a hyper-space feature.  Mathematically, an encoding block $e_\mathrm{FF}^{\mu}$ is used to encode a dynamic coordinate $(\mathbf{x}_\mathrm{d},t)$:

\begin{equation}
  \mathbf{F}_\mathrm{FF} = e_\mathrm{FF}^{\mu}\left( {\mathbf{x}_\mathrm{d},t} \right)
  \label{eq:c-ff}
\end{equation}

The free-form dynamic image $\mu_\mathrm{FF}$ is reconstructed using both $\mathbf{F}_\mathrm{TP}$ and  $\mathbf{F}_\mathrm{FF}$:

\begin{equation}
  \mu_\mathrm{FF}\left( {\mathbf{x}_\mathrm{d},t} \right) = \phi_\mathrm{FF}^{\mu}\left( {\mathbf{F}_\mathrm{TP},\mathbf{F}_\mathrm{FF},\mu_\mathrm{AR}\left( \mathbf{x}_\mathrm{s} \right)} \right)
  \label{eq:mu-ff}
\end{equation}

where $\phi_\mathrm{FF}^{\mu}$ is a prediction block for free-form dynamic image in the INR-based Reconstruction Module. Here, we incorporate an analytical reconstruction image $\mu_\mathrm{AR}$ generated by the Analytical Reconstruction Module to enhance the high-frequency detail representation. $\mu_\mathrm{AR}$ is aligned to the intermediate static coordinate space ($\mathbf{x}_\mathrm{s}$) through motion correction derived from DVFs in the Bidirectional DVF Estimation Module. Details about the analytical reconstruction module will be explained in Section \ref{sec:methods-modules-ar}.

The free-form dynamic image reconstruction  $\mu_\mathrm{FF}$ combines the strong topology-preserving foundation from $\mathbf{F}_\mathrm{TP}$, the hyper-space feature expansion via $\mathbf{F}_\mathrm{FF}$, and high-frequency enhancement from analytical reconstruction $\mu_\mathrm{AR}$. Although residual motion artifacts may persist in $\mu_\mathrm{AR}$, these artifacts can be effectively mitigated by $\mathbf{F}_\mathrm{FF}$. This combination enables $\mu_\mathrm{FF}$ to represent a wide range of dynamic CT images while effectively capturing rich high-frequency details. Consequently, $\mu_\mathrm{FF}$ serves as the final reconstruction output, while $\mu_\mathrm{TP}$ operates as a topology-preservation-based constraint. 

\begin{figure}[!t]
\centerline{\includegraphics[width=\columnwidth]{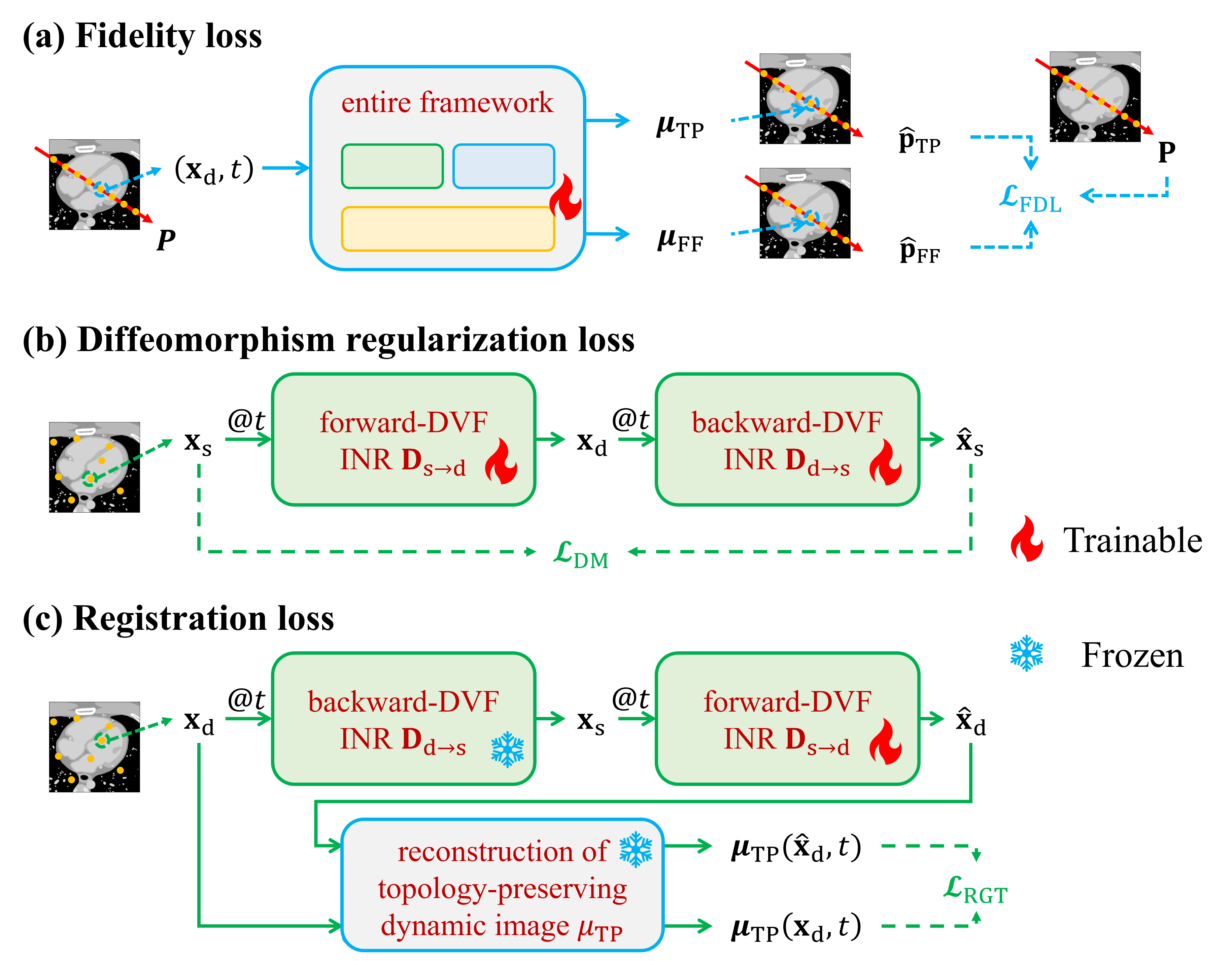}}
\caption{Illustration of the losses in the proposed framework.}
\label{fig-loss}
\end{figure}

With the above two reconstructions, we can define a data fidelity loss to directly supervise the predictions. Figure 2 illustrates the losses used in the proposed framework, where the data fidelity loss is in Figure \ref{fig-loss}-a. The data fidelity loss compares the post-log CT projections rendered from the predicted images with the ground-truth projections. Given the predicted projections $\hat{\mathbf{p}}$ rendered using \eqref{eq:math-proj-matvec} and the projections $\mathbf{p}$ acquired from a scan for n rays in a randomly sampled batch, our fidelity loss is defined using a Fourier domain L1-norm distance ($\rho^\mathrm{FD}$) calculated as:

\begin{equation}
  \rho^\mathrm{FD}\left( {\hat{\mathbf{p}},\mathbf{p}} \right) = {\frac{1}{n}\left\| {\mathrm{DFT}\left( \hat{\mathbf{p}} \right) - \mathrm{DFT}\left( \mathbf{p} \right)} \right\|}_{1}
  \label{eq:loss-fdl}
\end{equation}

The $\rho^\mathrm{FD}$ can help separate the different frequency components of errors, which we find helpful for the convergence of INR. Fidelity loss are computed for both $\mu_\mathrm{TP}$ and $\mu_\mathrm{FF}$, and their weighted sum is used as the total fidelity loss:

\begin{equation}
\begin{split}
  {\mathcal{L}_\mathrm{FDL}\left( {\mu_\mathrm{TP},\mu_\mathrm{FF}} \right)} 
&= \lambda_\mathrm{TP}\rho^\mathrm{FD}\left( {{\hat{\mathbf{p}}}_\mathrm{TP},\mathbf{p}} \right) \\
&\quad + \left( 1 - \lambda_\mathrm{TP} \right)\left. \rho^\mathrm{FD}\left( {{\hat{\mathbf{p}}}_\mathrm{FF},\mathbf{p}} \right) \right)
\end{split}
\label{eq:loss-fdl-total}
\end{equation}

where ${\hat{\mathbf{p}}}_\mathrm{TP}$ and ${\hat{\mathbf{p}}}_\mathrm{FF}$ are projections rendered from $\mu_\mathrm{TP}$ and $\mu_\mathrm{FF}$, respectively. $\lambda_\mathrm{TP} \in (0,1)$ balances the effect from the two terms. The loss on $\mu_\mathrm{FF}$ ensures that general dynamic variations can be accurately captured while the loss on $\mu_\mathrm{TP}$ strengthens the strong dimensional-reduction ability provided by topology-preserving feature. This dual supervision mechanism mitigates the framework’s over-reliance on either feature, and is critical for addressing the ill-poseness in dynamic CT reconstruction. The weight $\lambda_\mathrm{TP}$ can be tuned for different scenarios. For  dynamic objects known to be topology-invariant, $\lambda_\mathrm{TP}=0.9$ is recommended, for images significantly violating the topology-invariant assumption, $\lambda_\mathrm{TP}=0.1$ is suitable. A balanced value of $0.5$ is suggested for general cases.

In summary, the above dual-feature description of general dynamic objects maintains the topology-preserving ability and also extend to more general dynamic variations.

\subsubsection{Bidirectional DVF Estimation Module with diffeomorphism regularization}
\label{sec:methods-modules-bidvf}

As introduced in Section \ref{sec:methods-modules-dualfeature}, the proposed framework adopts backward-warping deformation to model the topology-preserving feature, where a backward-warping DVF is estimated using an INR. However, empirical analysis reveals that INR-based DVF estimation is prone to generate irregular and anatomically implausible results -- a critical limitation in ill-posed reconstruction such as nonperiodic dynamic CT. To address this issue, we encourage the estimated backward-warping DVF to be approximately diffeomorphic by introducing a complementary forward DVF INR and a diffeomorphism regularization loss enforcing inverse-consistency between these bidirectional DVFs. The resulted Bidirectional DVF Estimation Module is visualized in the green region of Figure 1.

\paragraph{Bidirectional DVF estimation}

In Bidirectional DVF Estimation Module, there are a backward DVF INR $\mathbf{D}_\mathrm{d\rightarrow s}$ that aligns the original dynamic coordinate space with the intermediate static space as in \eqref{eq:backward-warping}, and a forward DVF INR $\mathbf{D}_\mathrm{s\rightarrow d}$ for the inverse alignment as in \eqref{eq:forward-warping}.

Following the general architecture of INR in \eqref{eq:inr-general}, both backward and forward DVF INRs consist of an encoding block for spatial-temporal input, and a prediction block:

\begin{equation}
  \mathbf{D}_\mathrm{d\rightarrow s}\left( {\mathbf{x}_\mathrm{d},t} \right) = \phi_\mathrm{d\rightarrow s}^{\mathbf{D}}\left( {e_\mathrm{d\rightarrow s}^{\mathbf{D}}\left( {\mathbf{x}_\mathrm{d},t} \right)} \right)
  \label{eq:dvf-backward-structure}
\end{equation}

\begin{equation}
  \mathbf{D}_\mathrm{s\rightarrow d}\left( {\mathbf{x}_\mathrm{s},t} \right) = \phi_\mathrm{s\rightarrow d}^{\mathbf{D}}\left( {e_\mathrm{s\rightarrow d}^{\mathbf{D}}\left( {\mathbf{x}_\mathrm{s},t} \right)} \right)
  \label{eq:dvf-forward-structure}
\end{equation}

where $e_\mathrm{d\rightarrow s}^{\mathbf{D}}$ and $e_\mathrm{s\rightarrow d}^{\mathbf{D}}$ are the encoding blocks, and $\phi_\mathrm{d\rightarrow s}^{\mathbf{D}}$ and $\phi_\mathrm{s\rightarrow d}^{\mathbf{D}}$ are the prediction blocks.

\paragraph{Diffeomorphism regularization loss}

With the bidirectional DVF INRs, we design a diffeomorphism regularization loss inspired by GradICON \cite{tianGradICONApproximateDiffeomorphisms2023}. GradICON proposed a regularizer for approximate diffeomorphism in image registration task by penalizing the Jacobian of inverse consistency. It can offer faster convergence and improved regularity compared to directly penalizing the inverse consistency and other common regularizers.

In our framework, the inverse mapping is conducted by a first static-to-dynamic mapping followed by a second dynamic-to-static mapping:

\begin{equation}
\begin{matrix}
\hat{\mathbf{x}}_\mathrm{s} = \mathbf{x}_\mathrm{d} + \mathbf{D}_\mathrm{d\rightarrow s}\left( {\mathbf{x}_\mathrm{d},t} \right) \\
= \mathbf{x}_\mathrm{s} + \mathbf{D}_\mathrm{s\rightarrow d}\left( {\mathbf{x}_\mathrm{s},t} \right) + \mathbf{D}_\mathrm{d\rightarrow s}\left( {\mathbf{x}_\mathrm{s} + \mathbf{D}_\mathrm{s\rightarrow d}\left( {\mathbf{x}_\mathrm{s},t} \right),t} \right)
\end{matrix}
  \label{eq:dvf-icon-inverse}
\end{equation}

where $\hat{\mathbf{x}}_\mathrm{s}$ is the resulted estimation of cyclic-inversed static coordinate.

The diffeomorphism regularization loss $\mathcal{L}_{DM}$, as shown in Figure \ref{fig-loss}-b, penalizes the Jacobian of the inverse consistency:

\begin{equation}
 \begin{matrix}
\mathcal{L}_\mathrm{DM}\left( \mathbf{D}_\mathrm{s\rightarrow d},\mathbf{D}_\mathrm{d\rightarrow s} \right) = \left\| {\nabla\left\lbrack {\hat{\mathbf{x}}_\mathrm{s} - \mathbf{x}_\mathrm{s}} \right\rbrack} \right\|_{F}^{2} \\
= \left\| {\nabla\left\lbrack {\mathbf{D}_\mathrm{s\rightarrow d}\left( {\mathbf{x}_\mathrm{s},t} \right) + \mathbf{D}_\mathrm{d\rightarrow s}\left( {\mathbf{x}_\mathrm{s} + \mathbf{D}_\mathrm{s\rightarrow d}\left( {\mathbf{x}_\mathrm{s},t} \right),t} \right)} \right\rbrack} \right\|_{F}^{2}
\end{matrix}
  \label{eq:dvf-icon-jacobian}
\end{equation}

where, $\nabla$ denotes the Jacobian with repsect to $\mathbf{x}_\mathrm{s}$, and $\left\| \cdot \right\|_{F}$ represents the Frobenius norm. While theoretical computation of $\mathcal{L}_\mathrm{DM}$ involves integration across the entire 4D space, we approximate this by evaluation on a randomly sampled set of $\mathbf{x}_\mathrm{s}$ and $t$ each iteration. The Jacobian is implemented using PyTorch’s automatic differentiation package \texttt{torch.autograd}.

The proposed diffeomorphism regularization ensures the realism and reliability of the DVFs, which in turn improves the overall accuracy of the image reconstruction. The influence of diffeomorphism regularization is demonstrated in experiments in Section \ref{sec:results}.

\paragraph{Registration loss for forward-warping DVF}

The validity of previous diffeomorphism regularization relies on the normal function of both forward DVF INR $\mathbf{D}_\mathrm{s\rightarrow d}$ and backward DVF INR $\mathbf{D}_\mathrm{d\rightarrow s}$. While backward DVF INR $\mathbf{D}_\mathrm{d\rightarrow s}$ is inherently guided by the constraints from deformation-based image modeling, no objective exists for forward-warping DVF $\mathbf{D}_\mathrm{s\rightarrow d}$ to ensure the estimation of an actual forward-warping DVF. To address this gap, we introduce a registration loss for $\mathbf{D}_\mathrm{s\rightarrow d}$, explicitly training it to perform a static-to-dynamic image registration. Figure \ref{fig-loss}-c illustrates the registration lossforward-warping DVF $\mathbf{D}_\mathrm{s\rightarrow d}$.

The registration loss measures the discrepancy between the registered moving image values and the ground-truth fixed values. In our INR-based framework, both moving image and fixed image are represented by the proposed topology-preserving INR model that reconstructs pixel values from dynamic coordinates. Since the registration loss is computed for the forward-warping DVF INR $\mathbf{D}_\mathrm{s\rightarrow d}$, the “fixed image” corresponds to the INR-reconstructed values at ground-truth dynamic coordinates, while the “moving image” refers to values at estimated dynamic coordinates obtained by registering static input coordinates via $\mathbf{D}_\mathrm{s\rightarrow d}$. The registration task thus minimizes the discrepancy between image value evaluated at these two sets of coordinates.

Following the above definitions, the registration loss is implemented as follows. First, we sample a batch of ground-truth dynamic coordinates $\left\{ \left( \mathbf{x}_{\mathrm{d}}, t \right) \right\}$, and compute their corresponding image values $\left\{ \mu_\mathrm{TP}\left( {\mathbf{x}_\mathrm{d},t} \right) \right\}$ as the fixed image values. The static coordinates $\left\{ \mathbf{x}_\mathrm{s} \right\}$ are then derived from $\left\{ \mu_\mathrm{TP}\left( {\mathbf{x}_\mathrm{d},t} \right) \right\}$ via $\mathbf{D}_\mathrm{d\rightarrow s}$, serving as the input moving coordinates for registration. These static coordinates are warped to the estimated dynamic coordinates $\left\{ {\hat{\mathbf{x}}}_\mathrm{d} \right\}$ using $\mathbf{D}_\mathrm{s\rightarrow d}$. The moving image values $\left\{ \mu_\mathrm{TP}\left( \hat{\mathbf{x}}_\mathrm{d},t \right) \right\}$ are reconstructed at the registered coordinates. Finally, the registration loss $\mathcal{L}_\mathrm{RGT}$ is computed as the voxel-wise L1 distance between the registered moving image values $\left\{ \mu_\mathrm{TP}\left( \hat{\mathbf{x}}_\mathrm{d},t \right) \right\}$ and the ground-truth fixed image values $\left\{ \mu_\mathrm{TP}\left( {\mathbf{x}_\mathrm{d},t} \right) \right\}$:

\begin{equation}
 \begin{matrix}
\mathbf{x}_{\mathrm{s}} = \mathbf{x}_{\mathrm{d}} + \mathbf{D}_{\mathrm{d} \to \mathrm{s}} \left( \mathbf{x}_{\mathrm{d}}, t \right) \\
    \hat{\mathbf{x}}_{\mathrm{d}} = \mathbf{x}_{\mathrm{s}} + \mathbf{D}_{\mathrm{s} \to \mathrm{d}} \left( \mathbf{x}_{\mathrm{s}}, t \right) \\
    \mathcal{L}_{\mathrm{RGT}} \left( \mathbf{D}_{\mathrm{s} \to \mathrm{d}} \right) = \left\| \mu_{\mathrm{TP}} \left( \hat{\mathbf{x}}_{\mathrm{d}}, t \right) - \mu_{\mathrm{TP}} \left( \mathbf{x}_{\mathrm{d}}, t \right) \right\|_1
\end{matrix}
  \label{eq:dvf-registration}
\end{equation}

During registration loss computation, only the forward DVF INR $\mathbf{D}_\mathrm{s\rightarrow d}$ is trainable, while all other network parameters remain fixed. Equipped with this registration loss, forward DVF INR is able to accomplish the role in diffeomorphism regularization.

To summarize, the Bidirectional DVF Estimation Module introduces bidirectional DVF INRs for both forward and backward warping, and a diffeomorphism regularization mechanism. These bidirectional DVF INRs jointly form a cycle for diffeomorphism regularization while simultaneously fulfilling distinct roles in the reconstruction pipeline. Specifically, the backward DVF INR facilitates dynamic-to-static alignment, essential for the deformation-based modeling of topology-preserving feature within the INR-based Reconstruction Module (Section \ref{sec:methods-modules-dualfeature}). Conversely, the forward DVF INR provides critical static-to-PAR alignment utilized in the Analytical Reconstruction Module (Section \ref{sec:methods-modules-ar}). Therefore, beyond enhancing the anatomical realism and reliability of the estimated DVFs through diffeomorphism regularization, the Bidirectional DVF Estimation Module serves as a crucial connecting element that integrates the various components of the proposed framework.

\subsubsection{Detail enhancement via Analytical Reconstruction Module}
\label{sec:methods-modules-ar}

INR often suffers from prolonged training iterations and limited capacity to recover fine details during early training stages. To mitigate these limitations, we propose an Analytical Reconstruction Module, which leverages motion-compensated analytical reconstruction to extract high-frequency details from dynamic projections. The module generates an image $\mu_\mathrm{AR}$ that provides detail enhancement for the reconstruction of free-form dynamic image (see \eqref{eq:mu-ff} in Section \ref{sec:methods-modules-dualfeature}). Below, we detail the module's formulation.

Since the scanned object is dynamic, direct analytical reconstruction from dynamic projections introduces motion artifacts that degrade temporal resolution and obscure critical details. To address this challenge, we employ motion-compensated analytical reconstruction by aligning each projection frame's reconstruction to a shared static coordinate space according to the per-frame DVFs estimated by the preceding Bidirectional DVF Estimation Module introduced in Section \ref{sec:methods-modules-bidvf}, and summing up the deformed images to form the motion-compensated reconstruction. This approach effectively preserves temporal coherence and structural details. However, implementing this alignment strategy for hundreds or thousands of projection frames would be computationally expensive. Therefore, we segment projections into narrow angular groups corresponding to small temporal windows, and reconstruct these segments to generate Partial Angle Reconstruction (PAR) images, which preserve essential temporal information, mitigating motion artifact and retaining rich structural details. This PAR-based motion-compensated analytical reconstruction approach is conceptually similar to \cite{hahnReductionMotionArtifacts2016, hahnMotionCompensationRegion2017}, except that the DVFs are derived directly from our framework’s existing module rather than relying on explicit motion models.

We denote the PARs corresponding K non-overlapping projection groups as $\left\{ \mu_{k}^\mathrm{PAR} \right\},~k = 1,\ldots,K$, with corresponding centers of the time windows denoted by $\left\{ t_{k}^\mathrm{PAR} \right\},~k = 1,\ldots,K$. To estimate the value at the static coordinate $\mathbf{x}_\mathrm{s}$ of the motion-compensated reconstruction image, $\mathbf{x}_\mathrm{s}$ is deformed by the forward DVF INR $\mathbf{D}_\mathrm{s\rightarrow d}$ to the dynamic coordinate at each time window center $\left\{t_{k}^\mathrm{PAR}\right\}$:

\begin{equation}
  \mathbf{x}_{k}^\mathrm{PAR} = \mathbf{x}_\mathrm{s} + \mathbf{D}_\mathrm{s\rightarrow d}\left( {\mathbf{x}_\mathrm{s},t_{k}^\mathrm{PAR}} \right),~k = 1,\ldots,K
  \label{eq:ar-x-par}
\end{equation}

The motion-compensated analytical reconstruction $\mu_\mathrm{AR}$ is obtained by summing up values at the queried coordinates from all PARs:

\begin{equation}
  \mu_\mathrm{AR}\left( \mathbf{x}_\mathrm{s} \right) = {\sum\limits_{k = 1}^{K}{\mu_{k}^\mathrm{PAR}\left( \mathbf{x}_{k}^\mathrm{PAR} \right)}}
  \label{eq:5-ar-mu}
\end{equation}

The Analytical Reconstruction Module enhances the performance through several mechanisms. First, it provides implicit supervision for the Bidirectional DVF Estimation Module by establishing constraints on motion-compensated reconstruction. Second, the PAR-based reconstruction retains high-frequency details from dynamic projection data, enabling the subsequent INR-based Image Reconstruction Module to concentrate primarily on recovering low-frequency information, which can accelerate convergence while improving overall reconstruction quality. Third, although $\mu_\mathrm{AR}$ may exhibit residual motion artifacts due to limited temporal resolution of PAR, these artifacts can be effectively suppressed by the subsequent free-form feature $\mathbf{F}_\mathrm{FF}$ during the enhancement process. Finally, the module addresses the over-smoothing issue commonly observed in standard INR-based reconstruction approaches. The framework achieves this through PAR's customizable kernel, which allows simultaneous adjustment of texture characteristics in both the analytical reconstruction and subsequent INR-based free-form image reconstruction, thereby improving adaptability to clinical application requirements.

In conclusion, the proposed Analytical Reconstruction Module uses the existing forward DVF INR to extract high-frequency details from dynamic projection data, which can improves detail preservation in subsequent INR-based image reconstruction while simultaneously increasing the robustness of DVF estimation. Additionally, the module provides adjustable control over image textures, better accommodating the diverse requirements of clinical applications.

\subsection{Architecture of INRs in the framework}

The proposed framework utilizes several INRs with 3D and 4D input coordinates. As introduced in Section \ref{sec:methods-basic-inr}, all INRs in this work consist of an encoding block and a prediction block. The prediction block in all INRs shares a same structure, while the encoding block differs according to the dimension of input coordinates.

\subsubsection{The prediction block}

The prediction block is used in the DVF INRs ($\phi_\mathrm{d\rightarrow s}^{\mathbf{D}}$ in \eqref{eq:dvf-backward-structure} and $\phi_\mathrm{s\rightarrow d}^{\mathbf{D}}$ in \eqref{eq:dvf-forward-structure}) and the image INRs ($\phi_\mathrm{TP}^{\mu}$ in \eqref{eq:mu-tp} and $\phi_\mathrm{FF}^{\mu}$ in \eqref{eq:mu-ff}). They share a compact yet efficient architecture consisting of two sequential one-hidden-layer MLPs with a shortcut connection from the first MLP’s input to the second MLP’s input:

\begin{equation}
  \phi(\mathbf{z}) = \varphi_{2}(\mathbf{z}, \varphi_{1}(\mathbf{z}))
  \label{eq:inr-imp-pred}
\end{equation}

where $\phi$ is the prediction block, $\varphi_{1}$ and $\varphi_{2}$ are one-hidden-layer MLPs, and $\mathbf{z}$ is the input feature.

\subsubsection{The encoding block with 3D input}

The encoding block with 3D input is used in the image INRs ($e_\mathrm{TP}^{\mu}\left( \mathbf{x}_{s} \right)$ in \eqref{eq:c-tp}). We adopt the multi-resolution hashgrid encoding function \cite{mullerInstantNeuralGraphics2022} as the encoding block, which provides effective representation of complicated scenes with high memory and computational efficiency. This method works by mapping input coordinates to grid positions across multiple resolution levels, where each grid vertex retrieves a feature vector from a fixed-size table using a hash function. Features from neighboring vertices are interpolated based on the voxel position within the grid cell, and the resulting features from all resolution levels are concatenated to create a rich representation.

\subsubsection{The encoding block with 4D input}

\begin{figure}[!t]
\centerline{\includegraphics[width=\columnwidth]{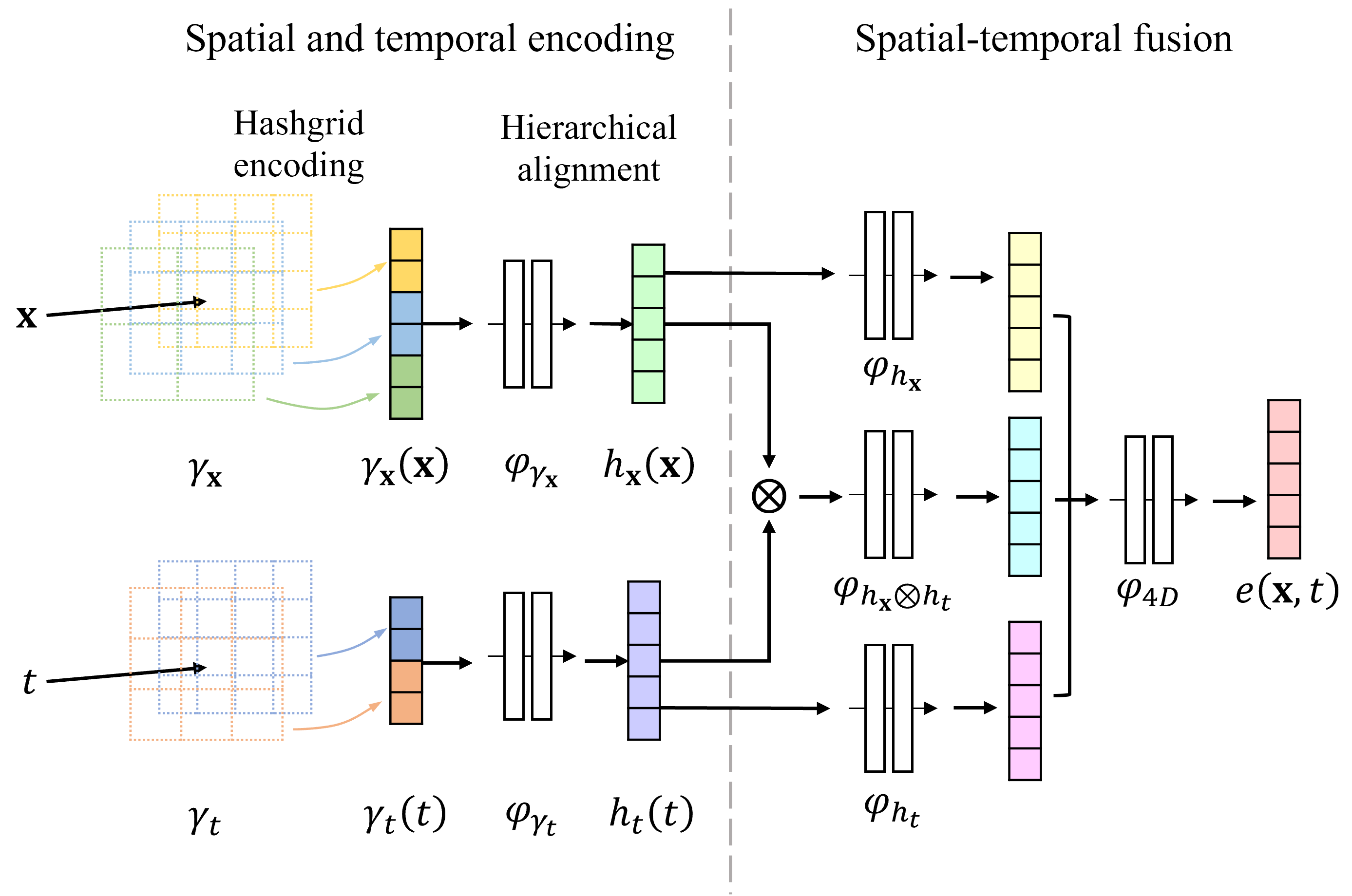}}
\caption{Illustration of the dimensional reduction structure for encoding block with 4D input.}
\label{fig-enc4d}
\end{figure}

The encoding blocks with 4D input is used in the DVF INRs ($e_\mathrm{d\rightarrow s}^{\mathbf{D}}\left( {\mathbf{x}_\mathrm{d},t} \right)$ in \eqref{eq:dvf-backward-structure} and $e_\mathrm{s\rightarrow d}^{\mathbf{D}}\left( {\mathbf{x}_\mathrm{s},t} \right)$ in \eqref{eq:dvf-forward-structure}), and the image INRs ($e_\mathrm{FF}^{\mu}\left( {\mathbf{x}_\mathrm{d},t} \right)$ in \eqref{eq:c-ff}). They share a dimensional-reduction structure that can both mitigate the ill-posed challenge and maintain the representational capacity. The proposed structure is illustrated in Figure \ref{fig-enc4d}. 

Dimensional-reduction is crucial for mitigating the challenges posed by measurement incompleteness in 4D space. In the proposed structure, dimensional-reduction is realized through decomposing 4D features into separate 3D spatial and 1D temporal components. This is motivated by the distinct characteristics of spatial and temporal variations within dynamic CT images and DVFs, where anatomical-related spatial information and motion-related temporal information often exhibit different change patterns and require independent treatment.

The proposed encoding block with 4D input consists of two steps: 1) individual spatial and temporal feature encoding, using a stack of hashgrid encoding and hierarchical alignment. 2) spatial-temporal fusion, derived from both the individual features and their cross-interaction.

\textbf{1) Spatial and temporal encoding:} to encode a paired input of 3D spatial coordinate $\mathbf{x}$ and a 1D temporal coordinate $t$, we use a sequential combination of multi-resolution hashgrid encoding and a hierarchical alignment MLP:

\begin{equation}
  h_{\mathbf{x}}\left( \mathbf{x} \right) = \varphi_{\gamma_{\mathbf{x}}}\left( {\gamma_{\mathbf{x}}\left( \mathbf{x} \right)} \right)
  \label{eq:enc4d-x}
\end{equation}

\begin{equation}
  h_{t}\left( t \right) = \varphi_{\gamma_{t}}\left( {\gamma_{t}\left( t \right)} \right)
  \label{eq:enc4d-t}
\end{equation}

Here, $\gamma_{\mathbf{x}}$ and $\gamma_{t}$ represent the 3D and 1D multi-resolution hashgrid encoding functions, while $\varphi_{\gamma_{\mathbf{x}}}$ and $\varphi_{\gamma_{t}}$ are single-hidden-layer MLPs. $h_{\mathbf{x}}$ and $h_{t}$ are the hierarchical-aligned encoded feature. The additional hierarchical alignment is to address the mismatch caused by difference in coordinate spaces and encoding functions, ensuring compatibility for the subsequent spatial-temporal fusion.

\textbf{2) Spatial-temporal fusion:} we model the spatial-temporal interaction as the combination of both individual spatial and temporal information, and their cross-interaction:

\begin{equation}
  e\left( {\mathbf{x},t} \right) = \varphi_\mathrm{4D}\begin{pmatrix}
{\varphi_{h_{\mathbf{x}}}\left( {h_{\mathbf{x}}\left( \mathbf{x} \right)} \right),} \\
{\varphi_{h_{t}}\left( {h_{t}(t)} \right),} \\
{\varphi_{h_{\mathbf{x}} \otimes h_{t}}\left( {h_{\mathbf{x}}\left( \mathbf{x} \right) \otimes h_{t}(t)} \right)}
\end{pmatrix}
  \label{eq:enc4d-fuse}
\end{equation}

In \eqref{eq:enc4d-fuse}, the hierarchical-aligned spatial and temporal encoded features ${h_{\mathbf{x}}\left( \mathbf{x} \right)}$ and ${h_{t}(t)}$ are proceeded through MLPs $\varphi_{h_{\mathbf{x}}}$ and $\varphi_{h_{t}}$ to model the individual information, respectively. The cross-interaction of ${h_{\mathbf{x}}\left( \mathbf{x} \right)}$ and ${h_{t}(t)}$ is also modeled by element-wise multiplication (denoted as $\otimes$ ) followed by an MLP $\varphi_{h_{\mathbf{x}} \otimes h_{t}}$. Finally, a fusion MLP $\varphi_\mathrm{4D}$ maps the concatenation of both individual and cross-interaction features into the final 4D encoded feature.

In conclusion, the proposed dimensional reduction structure for encoding block with 4D input is highlighted by hierarchical alignment during individual feature encoding, and the modeling of both individual and the cross-interaction features during spatial-temporal feature fusion. Compared to simple spatial-temporal decomposition strategies, such as concatenation or element-wise multiplication, the proposed approach can achieve both effective dimensional reduction and sufficient capacity of modeling complex spatial-temporal relationships, offering a more efficient decomposition method for ill-posed 4D reconstruction.

\subsection{Implementation details}

As described in the above sections, the proposed framework uses several losses, including the data fidelity loss $\mathcal{L}_\mathrm{FDL}$, the diffeomorphism regularization loss $\mathcal{L}_\mathrm{DM}$, and a forward DVF registration loss $\mathcal{L}_\mathrm{RGT}$. $\mathcal{L}_\mathrm{RGT}$ is used only for updating $\mathbf{D}_\mathrm{s\rightarrow d}$, with other components frozen. $\mathcal{L}_\mathrm{FDL}$ and $\mathcal{L}_\mathrm{DM}$ are used for updating the entire framework. The total loss is:

\begin{equation}
  \mathcal{L}_\mathrm{total} = \mathcal{L}_\mathrm{FDL} + \lambda_\mathrm{DM} \cdot \mathcal{L}_\mathrm{DM} + \lambda_\mathrm{RGT} \cdot \mathcal{L}_\mathrm{RGT}
  \label{eq:loss-total}
\end{equation}

where $\lambda_\mathrm{DM}=1.0$ and $\lambda_\mathrm{RGT}=0.1$ are recommended for first attempt.

Since our framework employs backward-warping deformation, it enables ray-based training. During each training iteration, $\mathcal{L}_\mathrm{FDL}$ is computed on a randomly sampled batch of rays, where  each ray is defined by a unique combination of detector pixel indices and view angles. The number of rays is 1024 by default. $\mathcal{L}_\mathrm{DM}$ and $\mathcal{L}_\mathrm{RGT}$ are computed separately on two sets of random coordinates. The number of sampled coordinates for $\mathcal{L}_\mathrm{DM}$ and $\mathcal{L}_\mathrm{RGT}$ is 1/4 of the number of points sampled during ray rendering in $\mathcal{L}_\mathrm{FDL}$.

We use Adam optimizer with a cosine annealing learning rate scheduler, to train the proposed INR framework. The initial learning rate for hashgrid encoders are $3e{-2}$, and for other components are $3e{-3}$.

The whole framework is implemented using PyTorch, with the hashgrid encoders implemented using tiny-cuda-nn (\underline{https://github.com/NVlabs/tiny-cuda-nn}
). For the configuration of hashgrid encoder, we set the base-resolution to 16, the log2-hashmap-size to 19, the n-levels to 16, and the n-features-per-level to 2. The per-level-scale is set to make the max resolution in the hashgrid about 2 times of the desired reconstruction resolution. All MLPs use a hidden layer size of 32.

\section{Experimental results}
\label{sec:results}

\subsection{Simulation Study 1: dynamic XCAT phantom}

In the \nth{1} simulation study, we evaluate the capability of our proposed method to reconstruct dynamic images driven by pure deformable motion, using projections of the dynamic XCAT phantom \cite{segars4DXCATPhantom2010}.

The XCAT program \cite{segars4DXCATPhantom2010} provides a digital human body phantom that models cardiac and respiratory movements through deformable transformations, representing an ideal test case for pure deformation-based dynamic image. For this simulation study, we generated a dynamic CT image sequence of the XCAT cardiac region covering a complete cardiac cycle, at a heart rate of 120 beats per minute (bpm) and a respiratory rate of 24 bpm.

The generated volume dimensions were 384×384×320. We simulated flat-panel cone-beam CT projections consisting of 540 views, each corresponding to a unique temporal frame, under two different gantry rotation speeds: 0.25 seconds per rotation (s/rot) and 0.5 s/rot. The faster gantry speed (0.25 s/rot, conducting 2 gantry rotations per cardiac cycle) approximates the speed limit in contemporary cardiac CT systems. The slower gantry speed (0.5 s/rot, conducting 1 gantry rotation per cardiac cycle) is typically insufficient for capturing cardiac motion at 120 bpm and was deliberately included to evaluate our method’s performance under extreme limited-angle conditions.

The comparison methods include conventional FDK and PICCS \cite{chenPriorImageConstrained2008}, and three variants of the proposed framework, as detailed in Table-\ref{tab:framework-variant}, that evaluated the contribution of the proposed diffeomorphism regularization and analytical reconstruction enhancement.

\begin{table}
\caption{Three variants of the proposed framework, evaluated in the experimental studies}
\setlength{\tabcolsep}{3pt}
\begin{tabular}{|>{\centering\arraybackslash}m{60pt}|>{\centering\arraybackslash}m{50pt}|>{\centering\arraybackslash}m{50pt}|>{\centering\arraybackslash}m{50pt}|}
\hline\hline
 & 
INR-base& 
INR-DM&
INR-DM-AR\\
\hline

DVF regularization&
L2 penalty&
diffeomorphism& 
diffeomorphism \\
\hline

analytical reconstruction enhancement&
$\times$&
$\times$& 
$\checkmark$ \\

\hline
\hline
\end{tabular}
\label{tab:framework-variant}
\end{table}

\begin{figure}[!t]
\centerline{\includegraphics[width=\columnwidth]{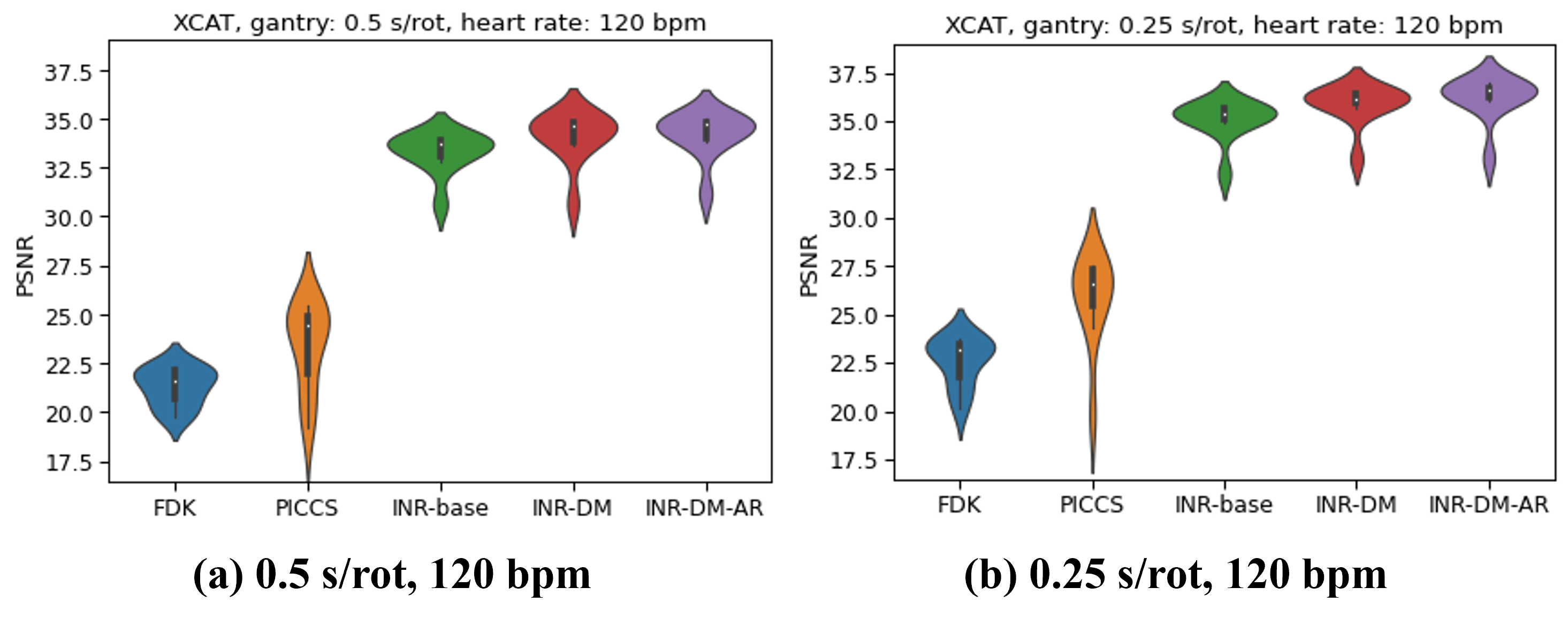}}
\caption{The violin-plot of PSNR counted on 10 phases of the XCAT phantom in the \nth{1} simulation study.}
\label{fig-psnr-xcat}
\end{figure}

For quantification evaluation, we sampled 10 ground-truth phases of XCAT-generated images, evenly distributed throughout the cardiac cycle. Figure \ref{fig-psnr-xcat} shows the distribution of per-phase PSNR for each method across both gantry rotation speed settings. All variants of the proposed framework significantly outperform FDK and PICCS. The inclusion of the proposed diffeomorphism regularization and analytical reconstruction enhancement each contributes to PSNR improvements.

\begin{figure}[!t]
\centerline{\includegraphics[width=\columnwidth]{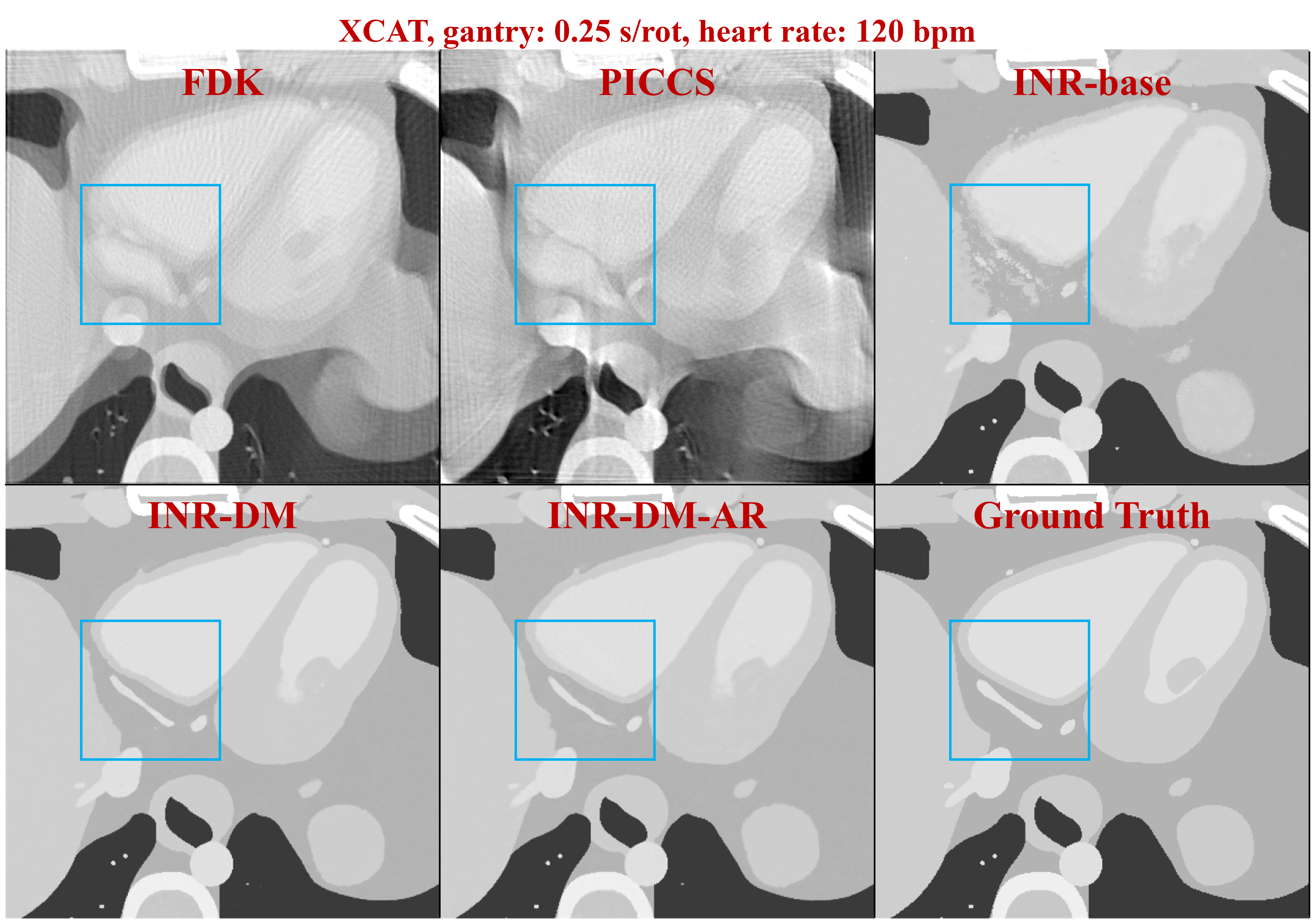}}
\caption{A reconstructed slice at the \nth{1} phase of the XCAT phantom in the \nth{1} simulation study, with the gantry rotation speed set to 0.25 s/rot, and the heart rate set to 120 bpm. The display window is [-600,300] HU.}
\label{fig-vis-xcat-1}
\end{figure}

\begin{figure}[!t]
\centerline{\includegraphics[width=\columnwidth]{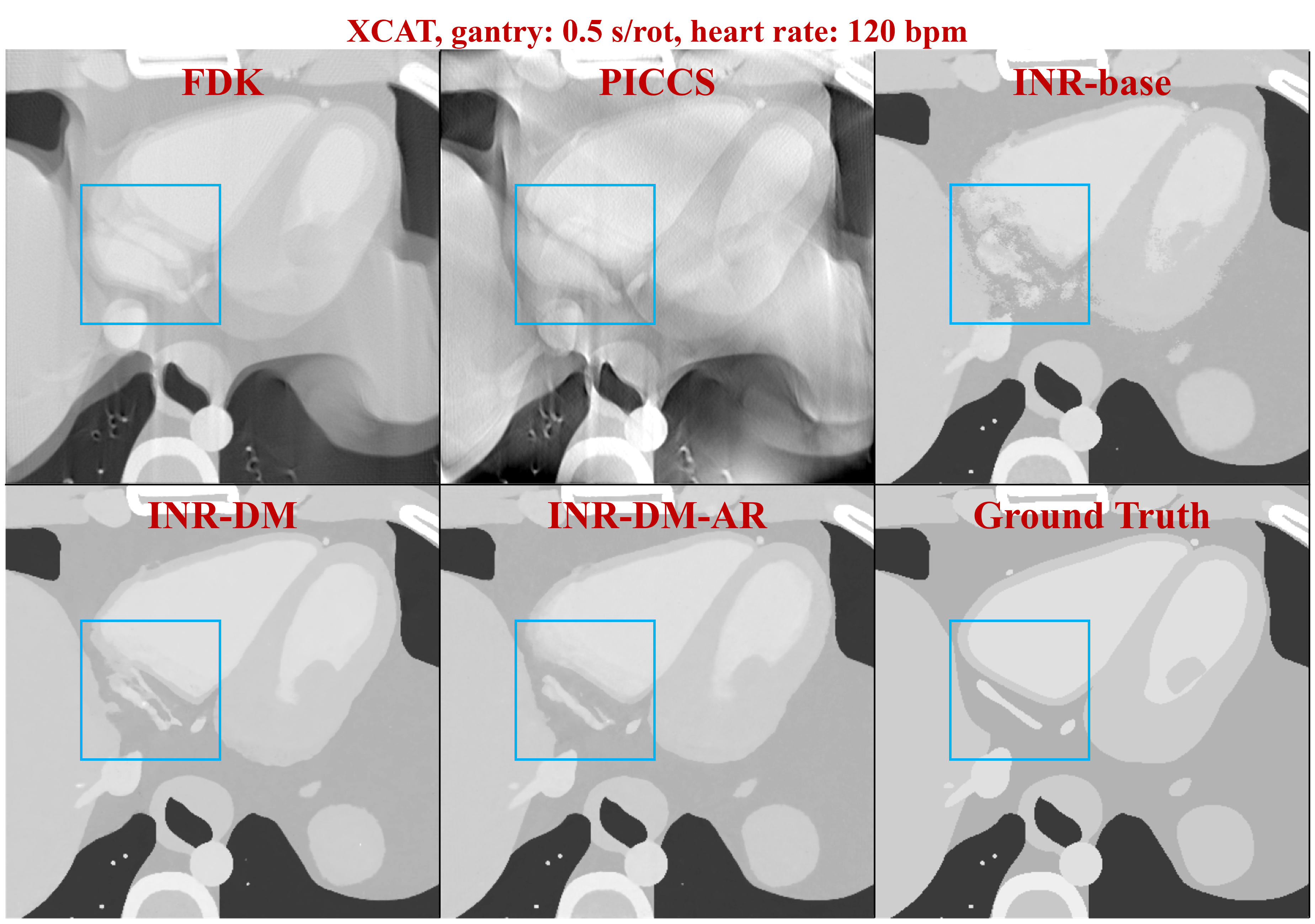}}
\caption{A reconstructed slice at the \nth{1} phase of the XCAT phantom in the \nth{1} simulation study, with the gantry rotation speed set to 0.5 s/rot, and the heart rate set to 120 bpm. The display window is [-600,300] HU.}
\label{fig-vis-xcat-2}
\end{figure}

Figures \ref{fig-vis-xcat-1} and \ref{fig-vis-xcat-2} present an example slice at the \nth{1} phase under different gantry rotation speed settings. In the less challenging scenario with gantry rotation speed of 0.25 s/rot (Figure \ref{fig-vis-xcat-1}), both FDK and PICCS produce images with severely blurred boundaries and geometrically distorted structures due to motion artifacts. In contrast, all variants of the proposed INR framework successfully generate images with clear structural boundaries and almost entirely removed motion artifacts. Notably, examination of a right coronary artery segment within the blue rectangular area demonstrates the framework's capability in reconstructing fine vascular structures despite motion. While INR-base coarsely reconstruct the structure, it exhibits limitations in reconstructing finer details due to poor DVF estimation, resulting in noisy boundaries throughout the image and fragmented vessel structures within the blue rectangular area. Both INR-DM and INR-DM-AR effectively reconstruct the entire image, mitigating motion artifacts and producing clearer boundaries compared to INR-base. Specifically, within the blue rectangular area, both variants reconstruct a continuous and well-defined vessel that closely approximates the ground-truth image. The superior performance of INR-DM and INR-DM-AR validates the effectiveness of diffeomorphism regularization. Under the gantry rotation speed of 0.25 s/rot, both INR-DM  and INR-DM-AR yield optimal result.

At the slower gantry rotation speed of 0.5 s/rot shown in Figure \ref{fig-vis-xcat-2}, all methods exhibit performance degradation to some extent. The FDK and PICCS reconstructions suffer from more pronounced motion artifacts. While all variants of the proposed INR framework still successfully reconstruct most structures with reduced motion artifacts, their performance within the blue rectangular area is noticeably compromised compared to the 0.25 s/rot case. The result of INR-base is especially worsened, presenting significantly noisier boundaries and completely fragmented, stretched vessel structures. In comparison, although INR-DM produces some vessel fragmentation, the extent is considerably less severe than INR-base, demonstrating the effectiveness of diffeomorphism regularization. INR-DM-AR maintains the highest fidelity, yielding vessel reconstructions that most closely approximate the ground-truth with minimal fragmentation, confirming the  effectiveness of  analytical reconstruction enhancement in mitigating motion artifacts and preserving fine anatomical structures.

\begin{figure}[!t]
\centerline{\includegraphics[width=\columnwidth]{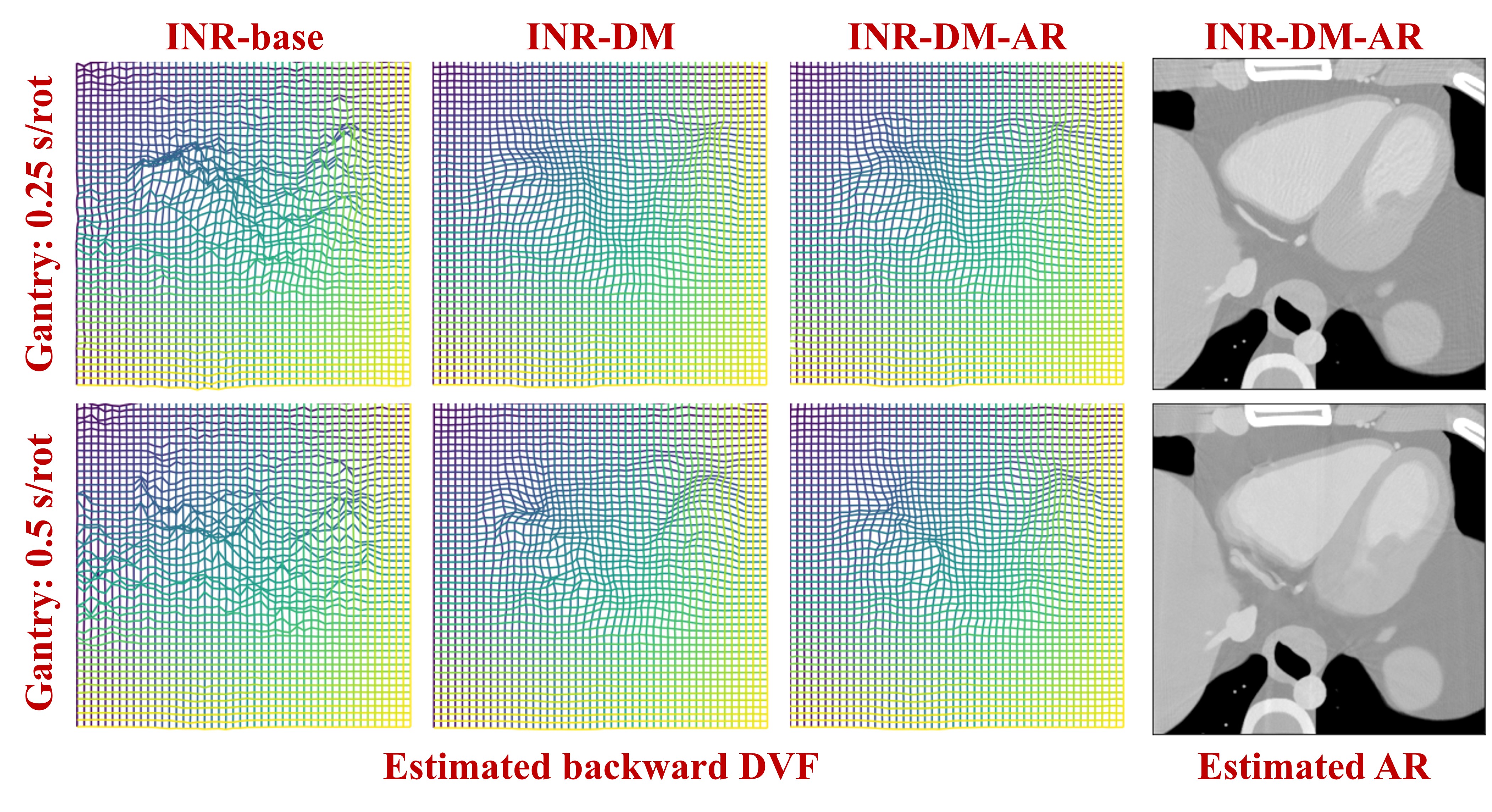}}
\caption{The visualization of estimated backward DVFs in different methods (the \nth{1} to \nth{3} columns), and estimated AR (analytical reconstruction) in INR-DM-AR (the \nth{4} column), in the \nth{1} simulation study. The gantry rotation speed is 0.5 s/rot (the \nth{1} row) and 0.25 s/rot (the \nth{2} row), respectively. The heart rate is 120 bpm. The display window is [-600,300] HU.}
\label{fig-dvf-xcat}
\end{figure}

Figure \ref{fig-dvf-xcat} visualizes the backward DVFs estimated by the three variations of the proposed INR framework, alongside the estimated analytical reconstruction in INR-DM-AR. The DVF estimated by INR-base appears notably noisy and unrealistic, whereas DVFs estimated by INR-DM and INR-DM-AR show improved smoothness and physiological plausibility. This observation suggests the critical importance of accurate DVF estimation in INR-based dynamic CT reconstruction. The analytical reconstruction in INR-DM-AR (Figure \ref{fig-dvf-xcat}) exhibits some streaking artifacts and residual motion artifacts compared to the final result of INR-DM-AR presented in Figure \ref{fig-vis-xcat-1} and Figure \ref{fig-vis-xcat-2}, but the underlying structures are remarkably similar. Considering the superior performance of INR-DM-AR over INR-DM in challenging ill-posed scenarios, it suggests that the proposed framework effectively leverages the initial motion compensation from the estimated analytical reconstruction, while simultaneously mitigating remaining artifacts through the free-form feature estimation, ultimately leading to optimal reconstruction quality, particularly in extreme limited-angle problems. 

In conclusion, the \nth{1} simulation study demonstrates the efficacy of the proposed framework for the reconstruction of pure deformation-based dynamic images.

\subsection{Simulation Study 2: clinical dynamic cardiac images}

In the \nth{2} simulation study, we access the method’s performance in reconstructing dynamic images containing both deformation-based movement and free-form variations, using projections generated from a clinical 20-phase cardiac CT image covering a complete cardiac cycle. This clinical dynamic image, acquired several years prior, contained distinct motion artifacts in each phase image due to the technical limitations of the acquisition system at that time. 

To ensure unique temporal frames for each projection view , we up-sampled the original 20-phase image to 2500 frames to match the number of simulated projection views. This up-sampling was achieved through VoxelMorph \cite{balakrishnanVoxelMorphLearningFramework2019}
registration and linear interpolation between adjacent frames. The motion artifacts present in the original 20-phase image were nonlinearly combined in each up-sampled frame, resulting in a dynamic image significantly affected by free-form variations in addition to the cardiac motion.

With these 2500 frames of cardiac images, we conducted simulations under two distinct heart rate settings: 120 bpm (2 gantry rotations per cardiac cycle) and 80 bpm (3 gantry rotations per cardiac cycle). The volume dimensions were 512×512×293. We simulated 2500-view flat-panel cone-beam CT projections, each view at a unique frame, with a gantry rotation speed of 0.25 s/rot. 

\begin{figure}[!t]
\centerline{\includegraphics[width=\columnwidth]{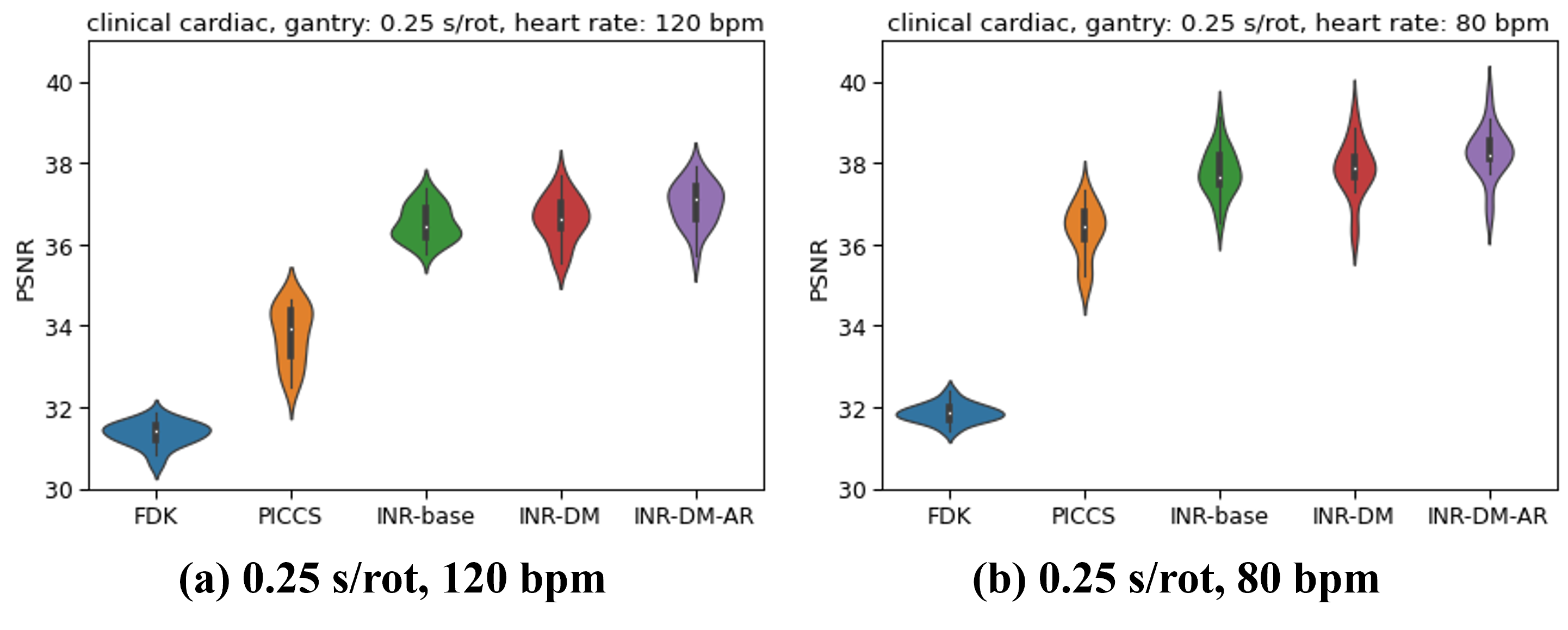}}
\caption{The violin-plot of PSNR counted on 20 phases of the cardiac image in the \nth{2} simulation study.}
\label{fig-psnr-cardiac}
\end{figure}

For quantification evaluation, the original 20-phase image is used as the ground-truth reference. Figure \ref{fig-psnr-cardiac} illustrates the distribution of per-phase PSNR for each method under both heart rate settings. The proposed framework demonstrates substantial advantages over conventional FDK and PICCS. Among its variants, INR-DM-AR achieves the highest overall accuracy, with INR-DM outperforming INR-base.

\begin{figure}[!t]
\centerline{\includegraphics[width=\columnwidth]{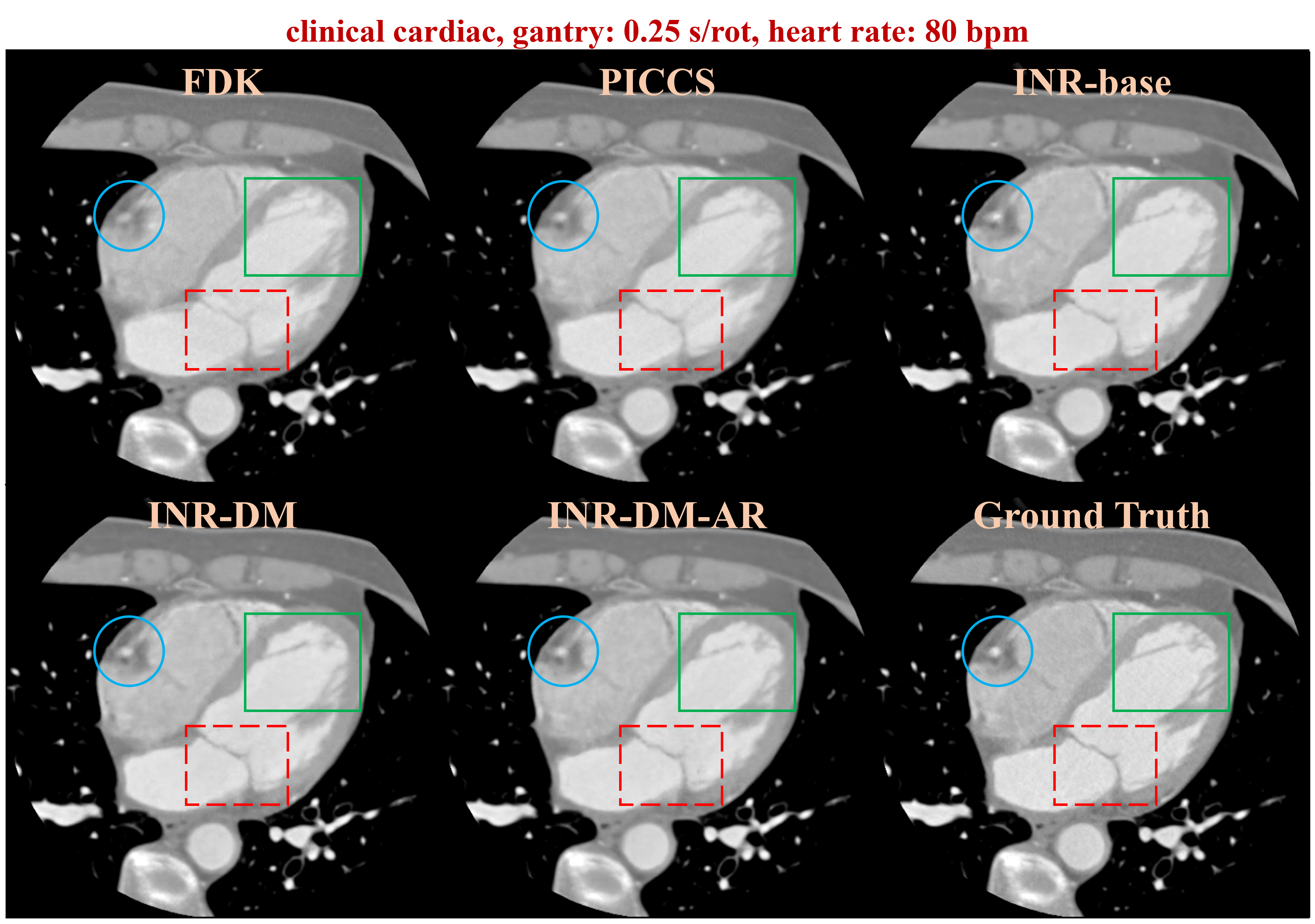}}
\caption{A reconstructed slice at the \nth{3} phase of the cardiac image in the \nth{2} simulation study, with the gantry rotation speed set to 0.25 s/rot, and the heart rate set to 80 bpm. The display window is [-768,512] HU.}
\label{fig-vis-cardiac-1}
\end{figure}

\begin{figure}[!t]
\centerline{\includegraphics[width=\columnwidth]{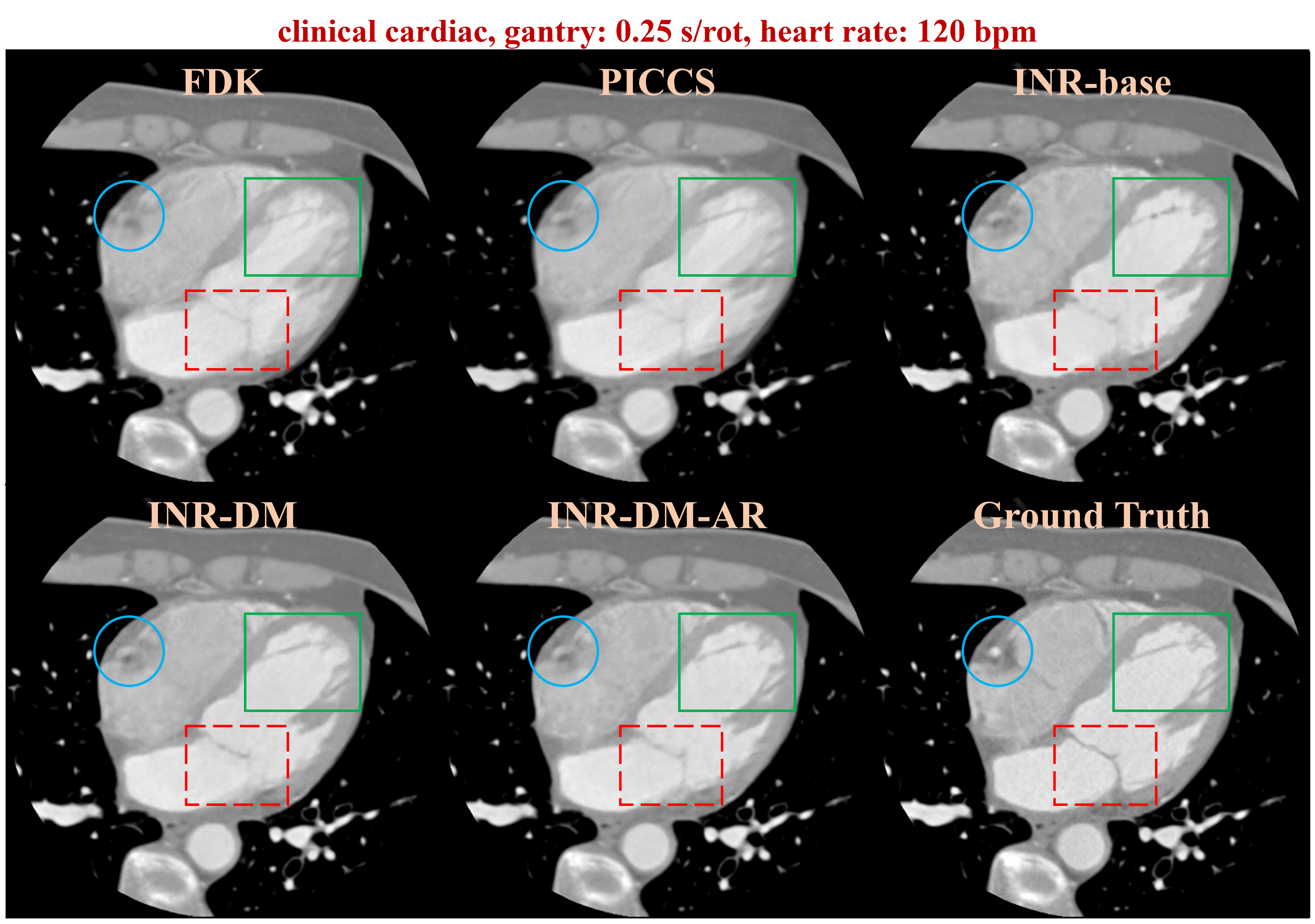}}
\caption{A reconstructed slice at the \nth{3} phase of the cardiac image in the \nth{2} simulation study, with the gantry rotation speed set to 0.25 s/rot, and the heart rate set to 120 bpm. The display window is [-768,512] HU.}
\label{fig-vis-cardiac-2}
\end{figure}

\begin{figure}[!t]
\centerline{\includegraphics[width=\columnwidth]{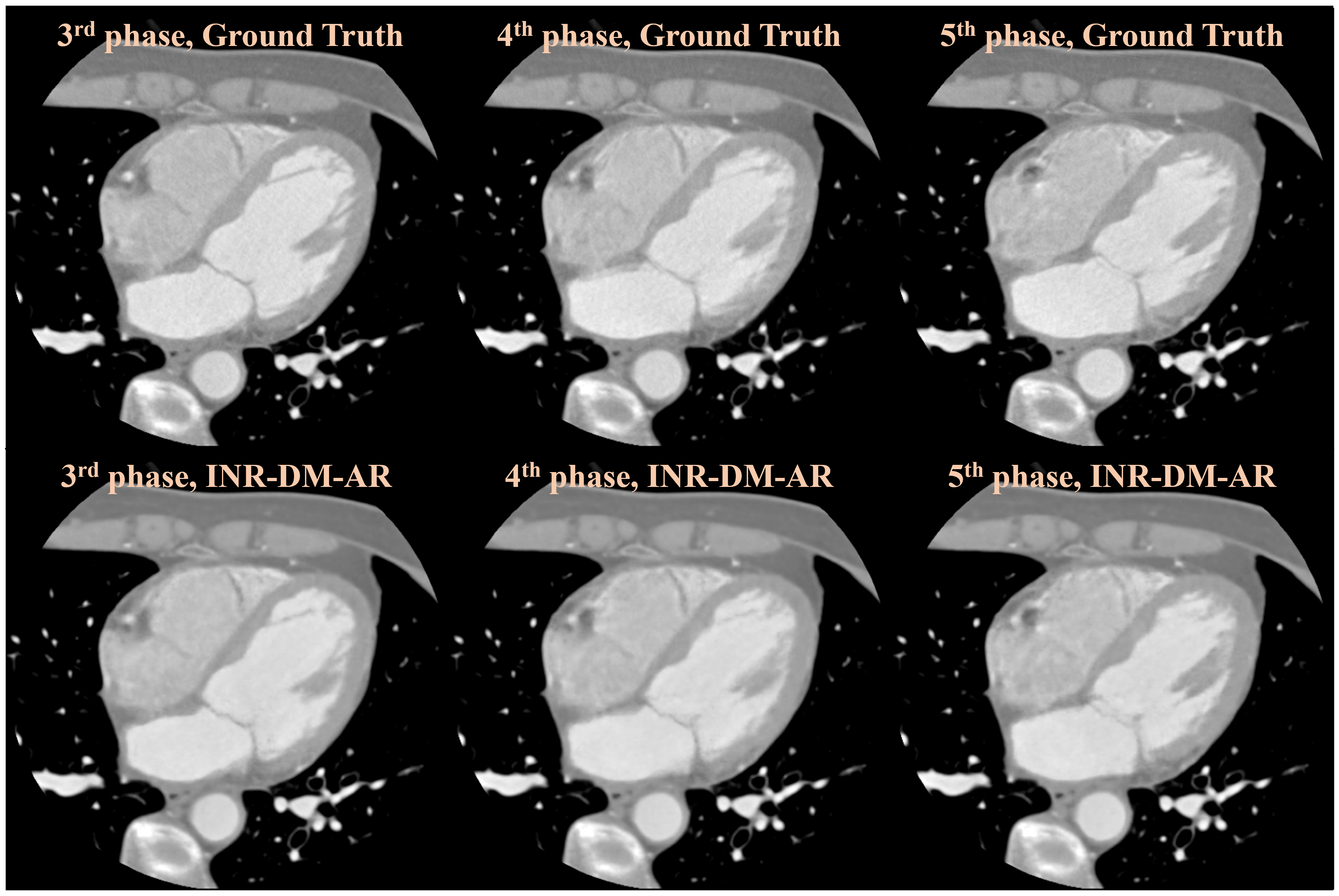}}
\caption{A reconstructed slice at the \nth{3}, \nth{4}, and \nth{5} phases of the cardiac image in the \nth{2} simulation study, with the gantry rotation speed set to 0.25 s/rot, and the heart rate set to 80 bpm. The display window is [-768,512] HU.}
\label{fig-vis-cardiac-adjacant}
\end{figure}

Figures \ref{fig-vis-cardiac-1} and \ref{fig-vis-cardiac-2} present an example slice from the \nth{10} phase  with heart rates of 120 and 80 bpm, respectively. The annotations in Figure \ref{fig-vis-cardiac-1} and \ref{fig-vis-cardiac-2} identify three structures: right coronary artery (blue circle), left ventricle (the green solid rectangle), and interface between left ventricle and left atrium (red dashed rectangle). The 120-bpm case (Figure \ref{fig-vis-cardiac-1}) represents the more challenging scenario. Both FDK and PICCS reconstructions exhibit visibly inferior quality, manifesting as arterial blurring, indistinct tissue boundaries, and pronounced ghosting artifacts in the left ventricle. Within our proposed framework, INR-DM and INR-DM-AR significantly outperform INR-base in boundary definition and structural detail recovery. Specifically, INR-DM-AR achieves optimal boundary clarity and ventricular detail preservation. However, all variants demonstrate smoothing effects in right coronary artery reconstruction. This limitation stems from inherent motion artifacts in the ground-truth reference data, where certain phases’ image already exhibit arterial blurring. Related to the evaluated \nth{3} phase here, The first row of Figure \ref{fig-vis-cardiac-adjacant} illustrates the adjacent \nth{3}, \nth{4}, and \nth{5} phases’ ground-truth image. The motion artifacts at the right coronary artery of the ground-truth image become progressively more severe, demonstrating strong phase-specific free-form variations in the ground-truth image. The phase-specific artifacts are then sampled using limited angular range during CT projection, resulting in a cross-phase mixture of artifacts in all methods’ reconstruction.

Figure \ref{fig-vis-cardiac-2} demonstrates improved performance across all methods at a heart rate of 80 bpm, particularly in reconstruction of the coronary artery (blue circle) and the region interface (red dashed rectangle). FDK and PICCS continue to exhibit ghosting and streak artifacts. The proposed INR framework variants generate clearer boundaries and structural details, with INR-DM-AR maintaining superior performance. This improved reconstruction quality at lower heart rates correlates with increased angular coverage of CT projection per-phase, critical for the data with phase-specific free-form variations. In specific, the second row of Figure \ref{fig-vis-cardiac-adjacant} illustrates the INR-DM-AR reconstruction of the adjacent \nth{3}, \nth{4}, and \nth{5} phases at a heart rate of 80 bpm. INR-DM-AR manages to reconstruct the phase-specific free-form variations to a certain degree.

\begin{figure}[!t]
\centerline{\includegraphics[width=\columnwidth]{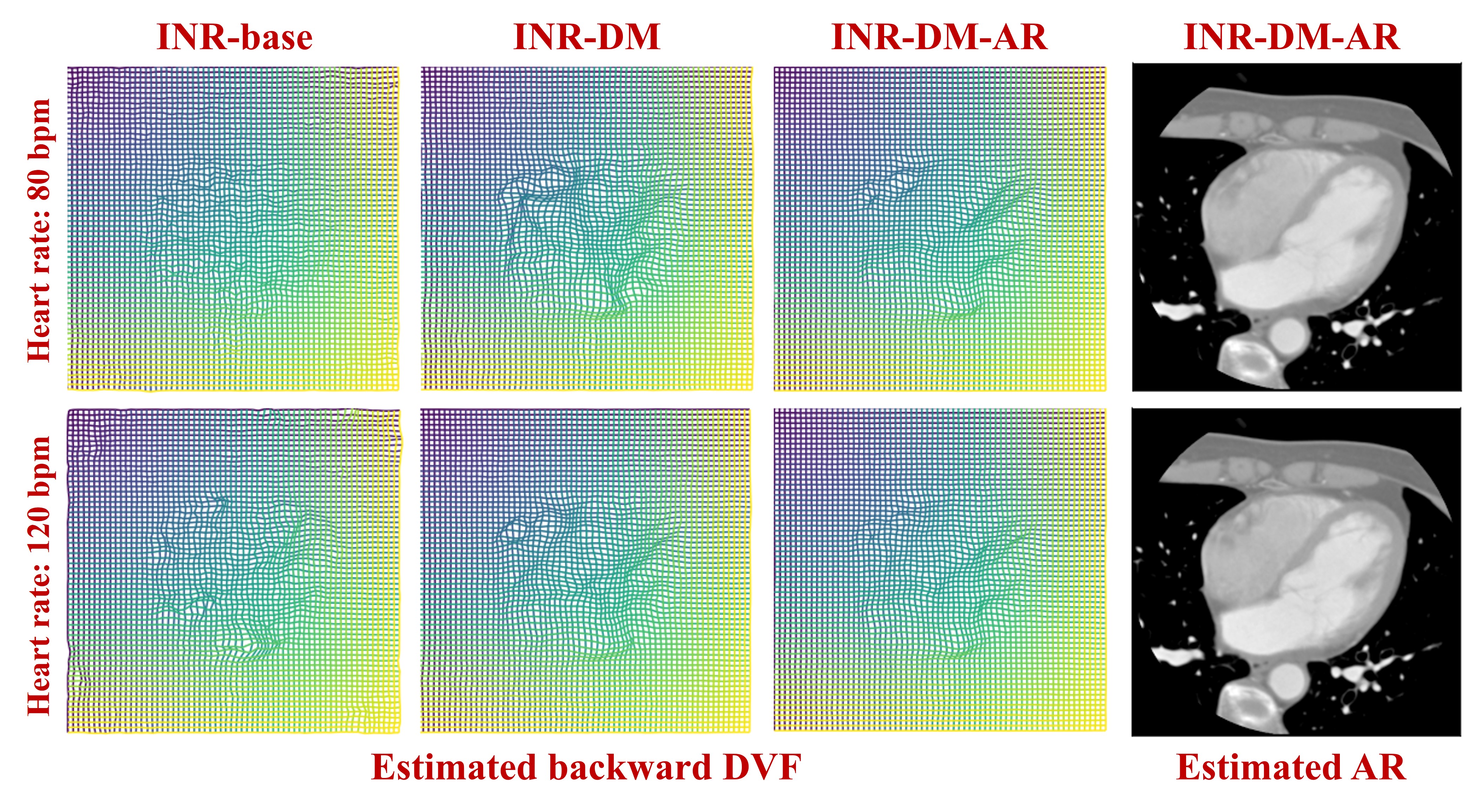}}
\caption{The visualization of estimated backward DVFs in different methods (the \nth{1} to \nth{3} columns), and estimated AR (analytical reconstruction) in INR-DM-AR (the \nth{4} column), in the \nth{2} simulation study. The heart rate is 80 bpm (the \nth{1} row) and 120 bpm (the \nth{2} row), respectively. The gantry rotation speed is 0.25 s/rot. The display window is [-768,512] HU.}
\label{fig-dvf-cardiac}
\end{figure}

Figure \ref{fig-dvf-cardiac} visualizes the backward DVFs estimated by the variants of the proposed INR framework, along with the analytical reconstruction estimated by INR-DM-AR. Consistent with findings from the \nth{1} simulation study, INR-DM and INR-DM-AR generate more physiologically plausible DVFs than INR-base, confirming the efficacy of the proposed diffeomorphism regularization. The analytical reconstruction exhibits greater blurring compared to the final output of INR-DM-AR, suggesting that analytical reconstruction becomes less informative for images dominated by free-form variations. This phenomenon arises because phase-specific free-form variations—unlike topology-preserving features that maintain cross-phase coherence—exhibit minimal correlation across PARs. Specifically, these variations manifest uniquely within each PAR, reducing the effectiveness of motion compensation in analytical reconstruction module. Nevertheless, INR-DM-AR partially recovers this phase-specific information through free-form feature estimation, addressing deficiencies inherent to both the topology-preserving feature and the analytical reconstruction, thereby achieving superior reconstruction quality.

In conclusion, the \nth{2} simulation study confirms the effectiveness of the proposed framework in reconstructing dynamic images influenced by both topology-preserving deformations and free-form variations.

\vspace{1cm}
\subsection{Practical Study 1: real projection data of a dynamic heart and lung phantom}

In the \nth{1} practical study, we evaluated the proposed method using a coronary CTA exam of a dynamic heart and lung phantom (PH-48, Kyoto Kagaku) scanned on a commercial 320-slice CT system (uCT 960+, United Imaging Healthcare). The acquisition employed an axial scanning with a gantry rotation speed of 0.25 s/rot. The reconstructed voxel size was 0.35 mm × 0.35 mm × 0.35 mm.

\begin{figure}[!t]
\centerline{\includegraphics[width=\columnwidth]{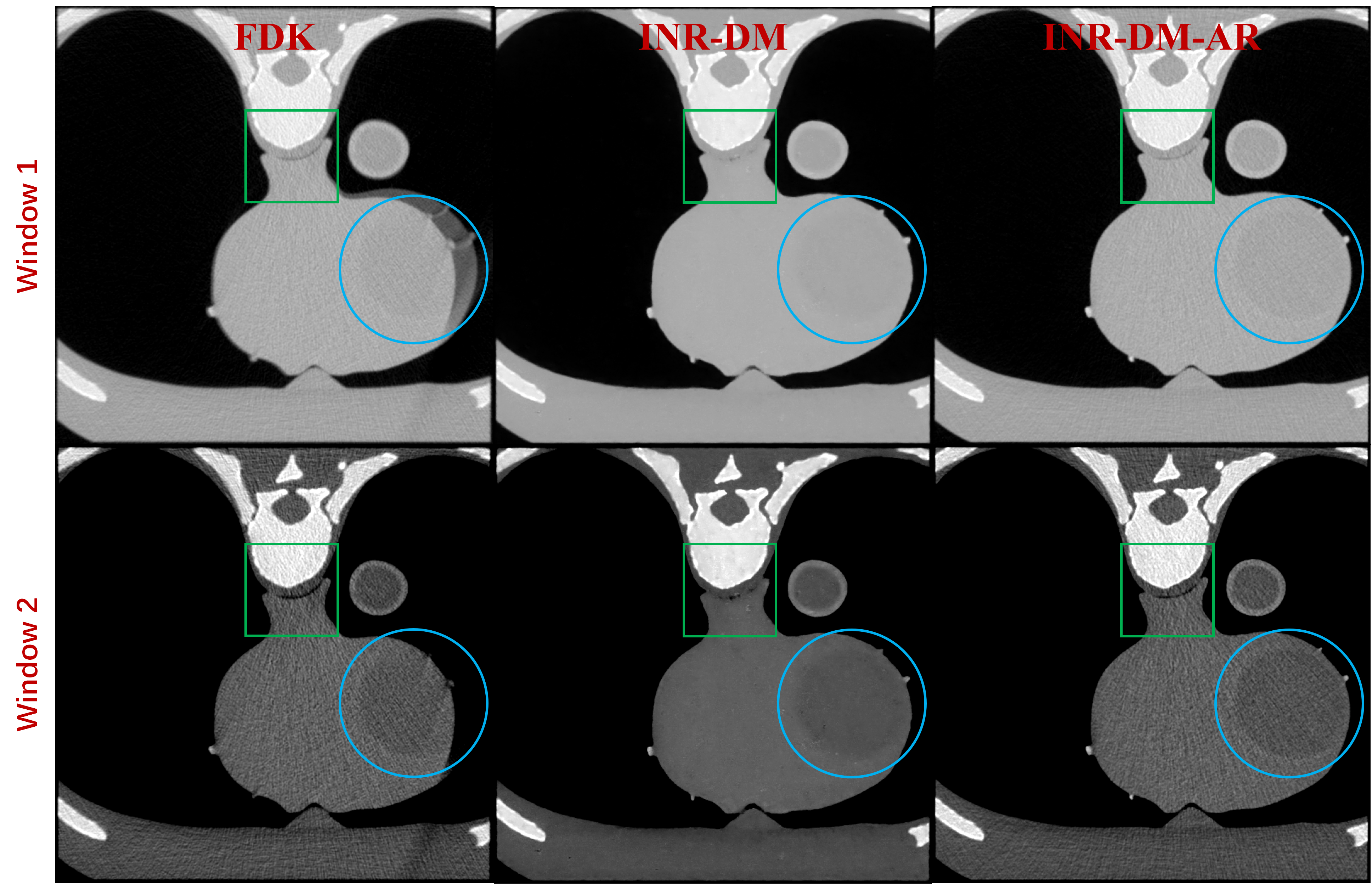}}
\caption{A reconstructed slice of the dynamic heart and lung phantom from an axial coronary CTA scan in the \nth{1} practical study, with the gantry rotation speed set to 0.25 s/rot. The display window is [-1024,512] HU for the \nth{1} row and [-250, 450] HU for the \nth{2} row.}
\label{fig-vis-phantom}
\end{figure}

The results are shown in Figure \ref{fig-vis-phantom}. The FDK reconstruction exhibits significant motion artifacts, whereas both INR-DM and INR-DM-AR successfully reconstruct the moving phantom with substantially reduced motion artifacts. In addition, INR-DM-AR demonstrates two distinct advantages over INR-DM. First, as evidenced in the green rectangle and blue circle regions, the tissue boundaries appear considerably sharper in INR-DM-AR compared to INR-DM, confirming the improved detail recovery achieved through analytical reconstruction enhancement. Second, INR-DM-AR produces images with noise characteristics and texture patterns similar to those of FDK reconstructions that employ identical reconstruction kernels as PARs. In contrast, INR-DM result exhibit slight over-smoothing. This validates the capability of analytical reconstruction enhancement to modulate image properties by adjustment of the PAR reconstruction kernels.

\subsection{Practical Study 2: retrospective real projection data of a patient}

In the \nth{2} practical study, we investigate the method’s capability to eliminate nonperiodic motion artifacts in a conventional CT scan. We utilized a retrospective arterial phase dataset of a multi-phase abdominal enhanced CT exam using a commercial 128-slice CT system (uCT 760, United Imaging Healthcare). The acquisition employed a helical scanning with a pitch ratio of approximately 1 and a gantry rotation speed of 0.5 s/rot. The reconstructed voxel size was 0.68 mm × 0.68 mm × 1.25 mm. Although the scan primarily targeted the abdomen region, it also captured a portion of the thoracic cavity. We therefore applied our method to reconstruct the thoracic region and compare the result with the reference reconstruction generated by the CT system’s standard algorithms. 

\begin{figure}[!t]
\centerline{\includegraphics[width=\columnwidth]{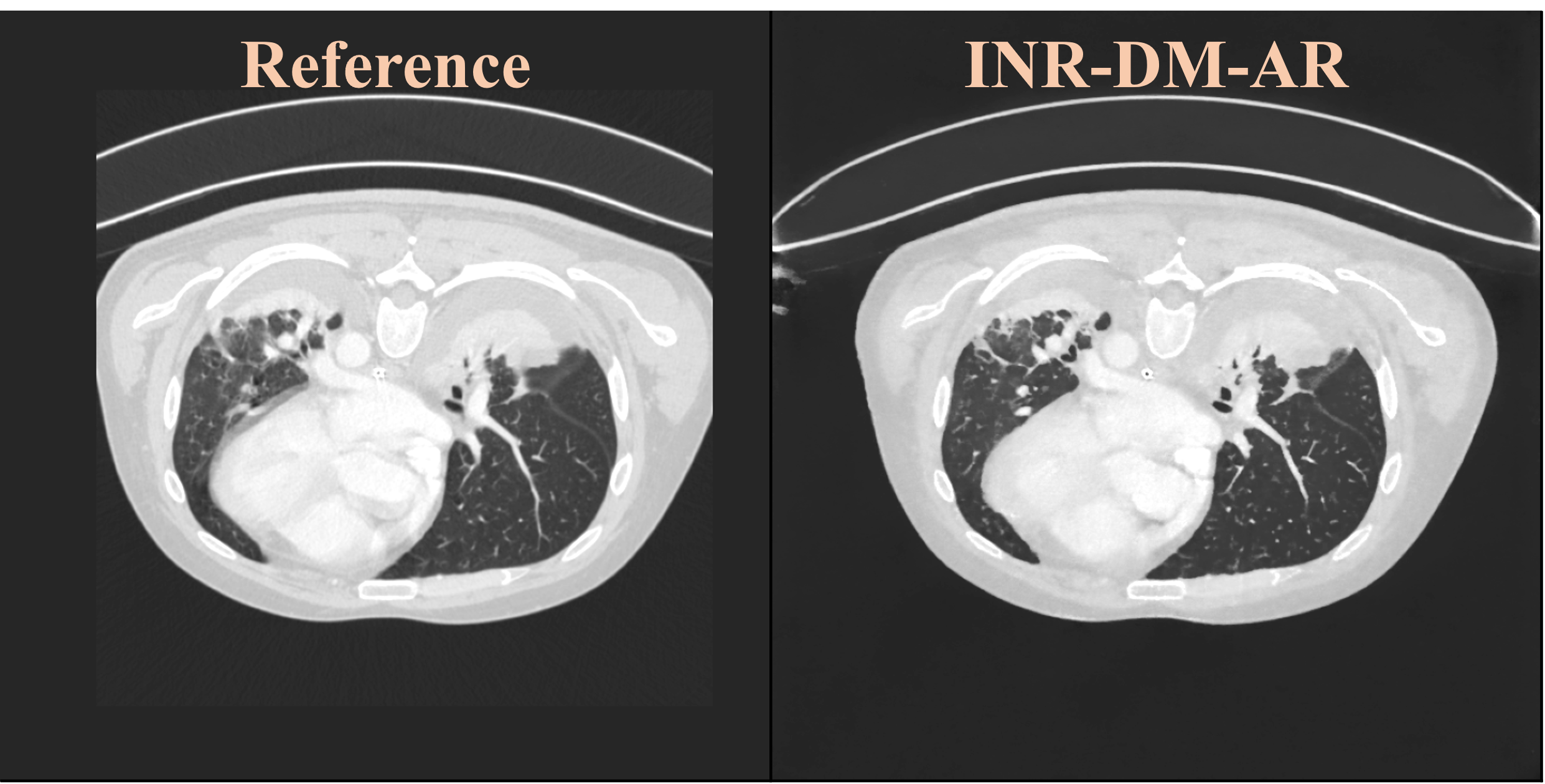}}
\caption{A reconstructed slice of the thoracic region from a retrospective helical abdominal CT scan in the \nth{2} practical study, with the gantry rotation speed set to 0.5 s/rot. The display window is [-1250,250] HU.}
\label{fig-vis-msct}
\end{figure}

Figure \ref{fig-vis-msct} illustrates the results. It is worth noting that the scan protocol employed—characterized by relatively slow gantry rotation speed and helical pitch ratio—was for abdominal imaging and is sub-optimal for cardiac applications. Consequently, the reference image exhibits pronounced motion artifacts induced by cardiac and respiratory movements, manifesting as blurring, smearing and double-contouring artifacts throughout the cardiac and pulmonary regions. In contrast, the proposed framework successfully reconstructs a clear dynamic image with effectively elimination of motion artifacts. The reconstruction exhibits sharp structures in pulmonary regions and more distinct region boundaries in the cardiac region. These results demonstrate the robust performance of the proposed framework in addressing nonperiodic motion.

\section{Conclusion and discussion}
\label{sec:conclusion}

In this study, we presented a novel INR-based framework for nonperiodic dynamic CT reconstruction that addresses several critical limitations of existing methods. Our framework incorporates a description of general dynamic object with topology-preserving and free-form features, backward-warping deformation modeling, diffeomorphism DVF regularization for anatomical plausibility, motion-compensated analytical reconstruction for detail enhancement, and dimensional-reduction design for efficient 4D coordinate encoding.

We conducted comprehensive validation through simulations and practical studies, covering digital and physical phantoms as well as retrospective real patient data. Our method achieves superior reconstruction quality compared to existing techniques in challenging limited-angle, nonperiodic motion scenarios. The backward-warping approach significantly reduces computational demands while maintaining high-resolution capabilities. The diffeomorphism regularization successfully balances complexity and anatomical realism in the estimated DVF. The motion-compensated analytical reconstruction effectively enhances fine details and controls image textures without requiring additional pre-scans, resolving the challenge of detail recovery in INR-based dynamic CT reconstruction.

Despite these advances, several limitations and opportunities for future work remain. While our approach reduces computational demands compared to forward-warping methods, the iterative optimization process remains computationally intensive for high-resolution applications. Potential solutions include advanced INR-based or Gaussian-splatting-based \cite{kerbl3DGaussianSplatting2023} models, more efficient diffeomorphism regularization, and hybrid approaches with data-driven methods. Additionally, DVF estimation methods could be improved to handle rapid object movement relative to gantry rotation speed, potentially incorporating data-driven or patient-specific knowledge.

In conclusion, our framework represents a significant step forward in addressing the fundamental challenges of nonperiodic dynamic CT reconstruction. By overcoming the computational limitations of forward-warping methods, ensuring anatomically plausible DVF estimation, and effectively preserving fine details without external historical scans, our approach enables more accurate and efficient reconstruction of dynamic CT images. Our method has several potential clinical applications. The first is nonperiodic dynamic CT reconstruction, such as one-beat cardiac imaging. The second is the generation of high-temporal-resolution cinematic image sequences that can be used for functional imaging. The third is to remove motion artifacts in normal CT scans to avoid repeated scans for patients with failed breath-holding, peristalsis or other movements.

\bibliographystyle{IEEEtran}
\bibliography{citations}

\end{document}